\documentclass[journal]{IEEEtran}

\ifCLASSINFOpdf

\else

\fi

\usepackage{amsmath,amsthm,amssymb,amsfonts}
\usepackage{graphicx}
\usepackage{bbding}

\newcommand{\ie}{\textit{i}.\textit{e}., }
\newcommand{\eg}{\textit{e}.\textit{g}., }
\newcommand{\etal}{\textit{et} \textit{al}.}

\usepackage{graphicx}
\usepackage{multirow}
\usepackage{hyperref}
\hypersetup{
	colorlinks=true,
	linkcolor=black,
	filecolor=black,      
	urlcolor=red,
	citecolor=black,
}

\begin{document}
	%
	\title{Enhancing Weakly Supervised Multimodal Video Anomaly Detection through Text Guidance}
	%
	%
	%
	
	\author{Shengyang Sun, Jiashen Hua, Junyi Feng, Xiaojin Gong,~\IEEEmembership{Member,~IEEE}
		\thanks{Shengyang Sun is with the School of Computer Science and Technology, Hangzhou Dianzi University, Hangzhou, China. \\
			Jiashen Hua and Junyi Feng are with the Alibaba Cloud, Hangzhou, China. \\ 
			Xiaojin Gong is with the College of Information Science and Electronic Engineering, Zhejiang University, Hangzhou, China. \\
			Corresponding author: Xiaojin Gong. E-mail: gongxj@zju.edu.cn \\
		Accepted for publication in IEEE Transactions on Multimedia. © 2026 IEEE. Personal use of this material is permitted. For permission to reprint/republish, contact IEEE.}}

	\markboth{Journal of \LaTeX\ Class Files,~Vol.~14, No.~8, August~2024}%
	{Shell \MakeLowercase{\textit{et al.}}: Bare Demo of IEEEtran.cls for IEEE Journals}

	\maketitle
	
	\begin{abstract}
		In recent years, weakly supervised multimodal video anomaly detection, which leverages RGB, optical flow, and audio modalities, has garnered significant attention from researchers, emerging as a vital subfield within video anomaly detection. However, previous studies have inadequately explored the role of text modality in this domain. With the proliferation of large-scale text-annotated video datasets and the advent of video captioning models, obtaining text descriptions from videos has become increasingly feasible. Text modality, carrying explicit semantic information, can more accurately characterize events within videos and identify anomalies, thereby enhancing the model's detection capabilities and reducing false alarms. 
		However, text feature extraction challenges anomaly detection. Pre-trained large language models often struggle to effectively capture the nuances associated with anomalies, as their training is based on generalized datasets. Directly fine-tuning the text feature extractor is also challenging, as anomaly-related text descriptions are sparse. 
		Furthermore, due to the varying amounts of information carried by different modalities, issues such as modality redundancy and modality imbalance arise during feature fusion. 
		To address the challenges of text feature extraction and the issues of modality redundancy and imbalance, we propose a novel text-guided weakly supervised multimodal video anomaly detection framework. Specifically, we introduce an in-context learning based multi-stage text augmentation mechanism to generate high-quality anomaly text samples. These high-quality samples are then used to fine-tune the text feature extractor, aiming to obtain a more effective text feature extractor for anomaly detection.  
		Additionally, we present a multi-scale bottleneck Transformer fusion module to enhance multimodal integration, utilizing a set of reduced bottleneck tokens to progressively transmit compressed information across modalities, aiming to address the issues of modality redundancy and imbalance. Experimental results on large-scale datasets UCF-Crime and XD-Violence demonstrate that our proposed approach achieves state-of-the-art performance. This project is publicly available at~\href{https://shengyangsun.github.io/TGMVAD}{https://shengyangsun.github.io/TGMVAD}.
		
	\end{abstract}
	
	
	\IEEEpeerreviewmaketitle
	
	\begin{figure}[t]
		\centering
		\includegraphics[width=0.46\textwidth]{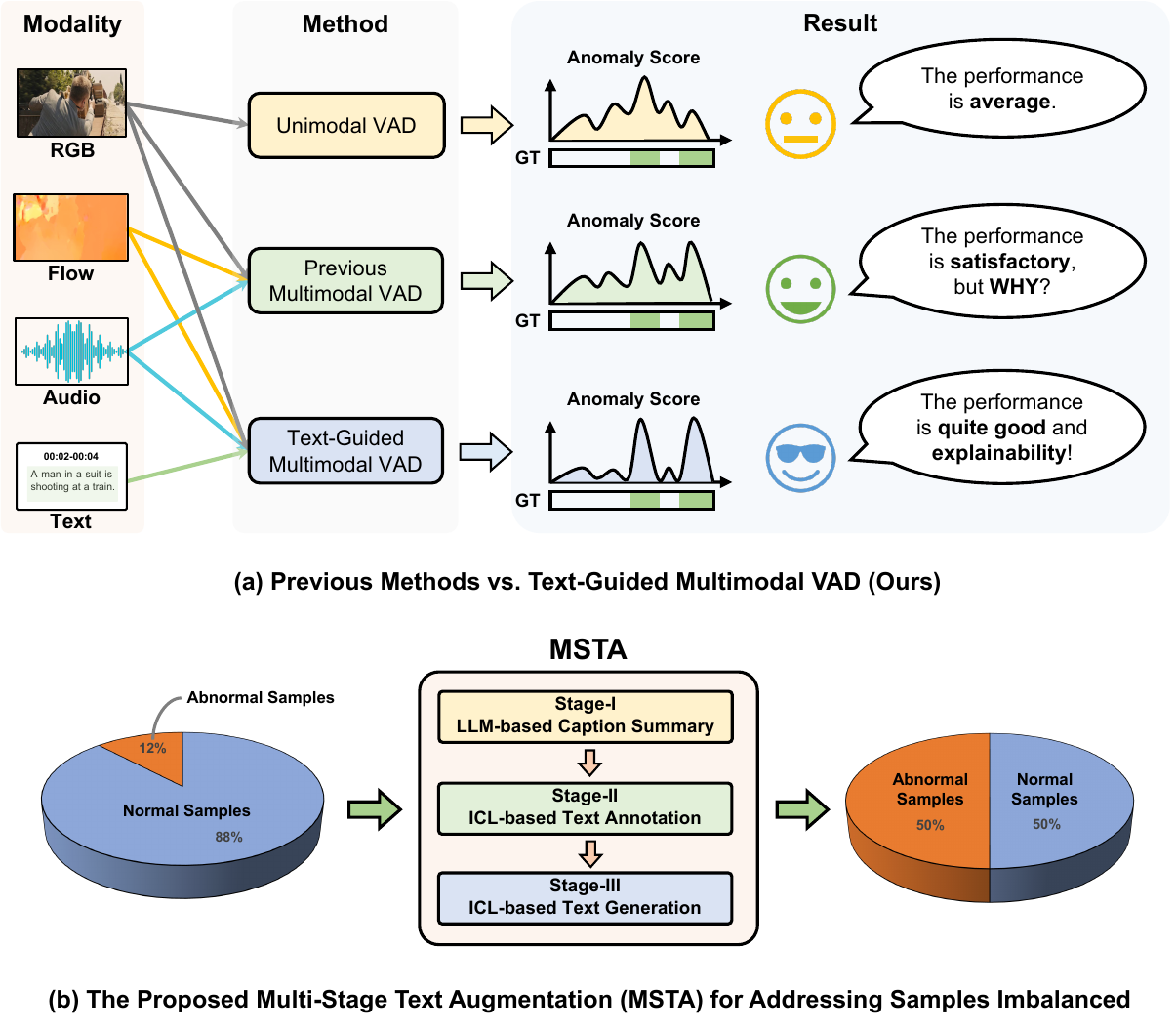} 
		\caption{(a) Previous methods for multimodal video anomaly detection (VAD) have demonstrated improved performance over unimodal VAD, however, they have inadequately explored the text modality and lack explainability. In contrast, we propose a text-guided multimodal VAD method that not only achieves superior performance but also provides enhanced explainability. (b) Using the UCF-Crime dataset as a case study, we observe that only 12\% of the text samples are classified as anomalies. This imbalance in the dataset introduces bias during the feature extractor fine-tuning. To address this issue, we propose the multi-stage text augmentation (MSTA) approach, which generates a greater number of high-quality anomaly samples to address the challenge of sample imbalance.}	
		\label{fig:index}
	\end{figure}
	
	\section{Introduction}
	\IEEEPARstart{V}{ideo} anomaly detection (VAD) has emerged as a pivotal task within computer vision and artificial intelligence, driven by the increasing prevalence of surveillance systems and the need for enhanced security measures. This task aims to identify unusual or unexpected events in video sequences, distinguishing them from normal patterns. Some earlier methods~\cite{landi2019anomaly,liu2019exploring} investigated training models in a fully supervised setting, which require extensive frame-level or snippet-level annotations for training, necessitating significant manual labeling efforts that can be both time-consuming and costly. While this approach can yield high accuracy in identifying anomalies, the resources required often limit its practical application. In contrast, the unsupervised methods~\cite{sun2022evidential,sun2023hierarchical} train the model solely on normal videos, recognizing outliers as anomalies during evaluation. Although these methods minimize labeling efforts, they pose challenges in real-world scenarios where obtaining a comprehensive set of normal samples is impractical. Furthermore, without exposure to abnormal samples, the model risks generating numerous false alarms. To bridge the gap between accuracy and labeling efficiency, the weakly supervised methods~\cite{sultani2018real,tian2021weakly,feng2021mist,chang2021contrastive,wu2022weakly,URDMU_zh,shi2023abnormal,sun2024multi} have gained attention, leveraging only video-level annotations for training. This setting significantly reduces labeling costs while still aiming to achieve robust performance, making it an increasingly popular choice for applications.

	In recent years, the field of weakly supervised multimodal video anomaly detection (WS-MVAD) has garnered increased attention for its potential to enhance anomaly detection performance through the integration of diverse data modalities. Unlike traditional unimodal approaches that primarily rely on RGB data~\cite{sultani2018real,tian2021weakly,feng2021mist,chang2021contrastive}, multimodal detection methods~\cite{pang2021violence,wei2022look,pang2022audiovisual,liu2022decouple,yu2022modality,sun2024multi} leverage complementary information from various sources, such as optical flow and audio, to achieve a more comprehensive understanding of video content. This multimodal strategy not only allows for a richer representation of contextual cues but also significantly improves the model's robustness and accuracy in identifying anomalies. For instance, while RGB data may struggle to differentiate between normal and anomalous behaviors in visually cluttered scenes, the inclusion of flow data can capture motion dynamics, and audio inputs can provide important contextual information regarding environmental sounds or conversations. By synergizing these modalities, weakly supervised multimodal approaches can mitigate the limitations associated with unimodal, thus leading to improved detection rates and reduced false alarms. This integrative approach represents a significant advancement in the quest for more effective video anomaly detection frameworks.
	Current multimodal video anomaly detection methods primarily employ simplistic approaches to fuse features from different modalities. Specifically, features from visual RGB, audio, or text modalities are typically fused via concatenation~\cite{wu2020not,wu2022weakly,chen2023tevad} or cross-attention mechanisms~\cite{wei2022look,liu2022decouple,yu2022modality}, and the fused features are then utilized for anomaly detection.
	For example, in the method proposed by Chen~\textit{et al.}~\cite{chen2023tevad}, the visual RGB modality features and text modality features are merged into fused features through concatenation for anomaly detection. Similarly, Yu~\textit{et al.}~\cite{yu2022modality} employ a cross-attention mechanism to fuse visual RGB and audio modality features. 
	
	These methods employ a straightforward feature fusion approach, making them relatively simple to implement and efficient. However, few current multimodal video anomaly detection methods have focused on the in-depth design of feature fusion modules that enable the fused features to carry more useful information. Besides, the role of text descriptions in multimodal-based video anomaly detection frameworks remains largely unexplored. Models that do not incorporate text modality tend to exhibit relatively inferior performance and lack interpretability, as shown in Fig.~\ref{fig:index}(a). Recently, advancements such as captioning models~\cite{lin2022swinbert,yang2023vid2seq,li2023blip} and the availability of large-scale text annotated datasets~\cite{yuan2024towards} have made text descriptions more accessible than ever, paving the way for their integration into anomaly detection task. The incorporation of text information not only facilitates a deeper contextual understanding of video content but also significantly enhances the model's performance by providing semantic insights that complement visual and auditory data. By leveraging text descriptions, models can improve their accuracy in identifying anomalies while reducing false positives through clearer definitions of expected behaviors. Moreover, these descriptions contribute to model explainability, enabling users to understand the underlying rationale behind detected anomalies, thereby fostering trust and facilitating actionable insights. As research progresses, harnessing the power of text descriptions in WS-MVAD promises to yield more robust and interpretable frameworks, capable of effectively addressing the complexities of real-world scenarios.
	
	However, WS-MVAD faces two significant challenges that hinder its effectiveness and reliability. One of the primary difficulties lies in the \textbf{1) modality redundancy and imbalance}~\cite{sun2024multi, nagrani2021attention, du2021improving}: each modality contains redundant information, which can introduce unintended semantic bias, \eg the noise introduced by background sounds within the audio modality. Besides, the information conveyed by one modality may significantly surpass that of another. If each modality is treated equally, it could result in a decline in detection performance. For example, the RGB typically conveys more information than the audio modality, resulting in a greater impact on detection performance. 
	Additionally, \textbf{2) text feature extraction challenges anomaly detection.} Pre-trained large language models often struggle to effectively capture the nuances associated with anomalies, as their training is based on generalized datasets. Consequently, employing these models directly for text feature extraction in anomaly detection may not yield remarkable performance. This limitation has prompted researchers to seek a text feature extractor fine-tuned with anomaly detection-specific data for the task of video anomaly detection. However, the imbalance between normal and anomalous text samples presents a significant challenge. Specifically, the rarity of anomalous events within the text descriptions of videos results in the predominance of samples detailing normal events, with only a small fraction describing anomalies. As shown in Fig.~\ref{fig:index}(b), the anomalous text samples in the UCF-Crime dataset account for merely 12\% of the total dataset. The overwhelming abundance of normal text samples during the fine-tuning of the feature extractor can introduce bias, potentially impairing the model's generalization capabilities and its effectiveness in accurately detecting anomalies.
	
	To address the aforementioned challenges in WS-MVAD, we propose a novel framework known as the text-guided multimodal video anomaly detection (TG-MVAD) framework. Central to our approach is a multi-stage text augmentation (MSTA) mechanism, designed to generate more high-quality abnormal samples that address the training bias during the fine-tuning of the text feature extractor. As illustrated in Fig.~\ref{fig:index}(b), by augmenting the training dataset with diverse and informative text descriptions, MSTA effectively mitigates the imbalance between normal and abnormal samples, facilitating a more comprehensive text extractor fine-tuning process, further obtaining a powerful text feature extractor for WS-MVAD. Furthermore, we introduce a multi-scale bottleneck Transformer (MSBT)-based fusion module~\cite{sun2024multi} that enhances inter-modality integration, addressing the modality redundancy and imbalance. This module utilizes a reduced set of bottleneck tokens to progressively transmit condensed information across modalities, thereby capturing intricate dependencies. Besides, our bottleneck token-based weighting scheme allows for adaptive feature weighting during the fusion process, optimizing the influence of each modality based on its relevance to the task at hand. Together, these innovations empower our framework to effectively fuse heterogeneous modalities, improve abnormal pattern recognition, and ultimately advance the field of WS-MVAD. Our contributions in this paper are summarized as follows:
	\begin{itemize}
		\item We introduce an in-context learning (ICL) based multi-stage text augmentation (MSTA) mechanism aimed at generating more high-quality abnormal text samples to counteract the bias of the text feature extractor fine-tuning. By enriching the training dataset with diverse and informative text descriptions, MSTA effectively reduces the imbalance between normal and abnormal samples, promoting a more effective fine-tuning process. 
		\item We propose a multi-scale bottleneck Transformer (MSBT) module that improves inter-modality integration. This module employs a reduced set of bottleneck tokens to progressively convey condensed information between modalities, effectively capturing complex dependencies.
		\item Experimental results on the large-scale datasets UCF-Crime and XD-Violence demonstrate that our proposed method achieves state-of-the-art results. Furthermore, the results of ablation studies indicate that our proposed MSTA significantly enhances the model's performance.
	\end{itemize}
	
	This paper is an extension of our previous conference work~\cite{sun2024multi}. We extended the initial version in several significant ways: 1) In this paper, we innovatively propose the MSTA mechanism, which generates more high-quality anomalous text samples intended for fine-tuning the text feature extractor. 2) Utilizing the high-quality text features extracted, we incorporate the text modality into the MSBT module for feature fusion. In contrast to previous works that exclusively employed RGB, optical flow, or audio modalities, the introduction of the text modality not only improves the model's performance but also gives explainability. 3) Furthermore, this work includes new experimental results on the large-scale UCF-Crime dataset, achieving a new state-of-the-art result, and further validating the effectiveness of our proposed method.
	
	\section{Related Work}
	
	\subsection{Video Anomaly Detection}
	Video anomaly detection (VAD) has garnered substantial research interest and can be categorized into three main areas: fully supervised methods~\cite{liu2019exploring, Acsintoae_2022_CVPR}, unsupervised approaches~\cite{sun2022evidential,sun2023hierarchical}, and weakly-supervised VAD~\cite{sultani2018real,tian2021weakly,feng2021mist,chang2021contrastive,wu2022weakly,URDMU_zh,shi2023abnormal,sun2023long,sun2024multi,sun2024event,sun2024tdsd}.
	
	The fully supervised VAD~\cite{landi2019anomaly,liu2019exploring} approach requires training a model with frame-level annotations that include precise bounding boxes for anomalies, an effort-intensive task due to the substantial labor involved in labeling the data.
	Landi~\etal~\cite{landi2019anomaly} enhance the surveillance videos in a dataset by adding spatial and temporal annotations, then train an anomaly detection model using bounding box supervision. Experiments demonstrate that a network trained with spatiotemporal tubes performs well.
	Liu~\etal~\cite{liu2019exploring} introduce an end-to-end trainable framework guided by anomalous regions, where they design a region loss under fully supervised learning to explicitly guide the network in identifying anomalous areas. Additionally, due to the challenges of deep networks and limited anomaly training data, the architecture incorporates a meta-learning module to mitigate overfitting.
	
	To mitigate the arduous task of annotating large quantities of samples, unsupervised methods~\cite{hasan2016learning,gong2019memorizing,sun2022evidential,sun2023hierarchical} concentrate on collecting only normal videos for model training. During the inference phase, these models then identify anomalies by detecting deviations from the learned normal patterns. For instance, reconstruction-based unsupervised techniques~\cite{hasan2016learning,gong2019memorizing,sun2023hierarchical} utilize autoencoders to encode normal samples into latent spaces. During inference, the model identifies poorly reconstructed samples as anomalies, as these deviations indicate a lack of fit with the learned normal patterns. 
	Hasan~\etal~\cite{hasan2016learning} employ a convolutional neural network (CNN)-based autoencoder to perform feature reconstruction, thereby labeling video frames with significant reconstruction errors as anomalies. This method relies on the powerful reconstruction capability of the autoencoder. However, a challenge is that the autoencoder can reconstruct both normal and anomalous video frames effectively, making it difficult to distinguish between normal and anomalous frames solely based on reconstruction errors.
	To address this issue, Gong~\etal~\cite{gong2019memorizing} introduce a memory bank module within the autoencoder-based architecture to store normal patterns. During the reconstruction process, the autoencoder utilizes weighted features retrieved from the memory bank, which alleviates the autoencoder's tendency to overly reconstruct anomalous frames.
	Sun and Gong~\cite{sun2023hierarchical} propose a hierarchical semantic contrastive learning method to develop a scene-aware model. Building on an autoencoder-based reconstruction framework, they introduce contrastive learning at both the scene level and object level. This approach encourages the encoded latent features to remain compact within the same semantic class, while also ensuring their separability across different classes.
	Distance-based unsupervised methods~\cite{Sabokrou2017Gaussian,ionescu2019detecting,sun2022evidential} employ Gaussian mixture models or one-class SVMs to establish decision boundaries, identifying data that deviates from these boundaries as anomalous. 
	Ionescu~\etal~\cite{ionescu2019detecting} propose a method that combines $K$-means clustering and support vector machines (SVMs) for anomaly detection in videos. They first apply $K$-means clustering to group normal samples and then use SVMs to construct the decision boundary for the normal class. Anomalies are identified by calculating the distance between each sample and the centroid of its corresponding cluster.
	Sun~\etal~\cite{sun2022evidential} propose a deep evidential reasoning network designed to encode evidence vectors and estimate uncertainty by learning evidence distributions and deriving beliefs from them. They introduce an unsupervised training strategy that minimizes an energy function based on the deep Gaussian mixture model (GMM). This method calculates the decision boundary and identifies frames that deviate from the normal pattern, classifying them as anomalies.
	However, because the training set lacks abnormal samples, the model often produces false alarms when encountering ambiguous normal data. To reduce the labor-intensive annotations and introduce abnormal samples during model training, weakly-supervised video anomaly detection (WS-VAD)~\cite{sultani2018real,tian2021weakly,feng2021mist,shi2023abnormal} addresses this challenge using multiple instance learning (MIL). By training the model with video-level annotations instead of detailed labels, this approach achieves excellent performance. 
	
	\subsection{Weakly Supervised Video Anomaly Detection}
	
	Weakly supervised learning has already been applied to several visual tasks, such as the weakly supervised point cloud segmentation~\cite{xue2024weakly}, weakly supervised semantic segmentation~\cite{zhang2023weakly}, and weakly supervised video anomaly detection~\cite{shi2023abnormal,chang2021contrastive}. For example, Xue~\etal~\cite{xue2024weakly} propose a local semantic embedding network that learns semantic propagation and local structural patterns. This network incorporates graph convolution-based dilation and erosion embedding modules to facilitate `inside-out' and `outside-in' pathways for semantic information dissemination. As a result, the proposed framework enables the mutual propagation of semantic information between the foreground and background.
	Zhang~\etal~\cite{zhang2023weakly} propose an alternate self-dual teaching (ASDT) framework, which is structured around a dual-teacher, single-student network architecture. The information interaction process between the different network branches is framed within the context of knowledge distillation. Given that the knowledge from the two teacher models may be noisy or faulty under weak supervision, they introduce a pulse-width wave-like selection signal to guide the distillation process. This signal helps prevent the student model from converging to trivial solutions caused by the imperfect knowledge of either teacher model.
	
	In video anomaly detection, weakly supervised learning-based methods utilize video-level labels for model training, and several approaches~\cite{zaheer2020claws,wu2024vadclip,chang2021contrastive,shi2023abnormal} have been investigated. For instance, 
	Zaheer\etal~\cite{zaheer2020claws} introduced a clustering loss during the model training process, aimed at improving the performance of video anomaly detection by optimizing the structure of the feature space. Specifically, this strategy minimizes the distance between feature clusters in normal videos to ensure that the features of normal videos are more tightly grouped in the feature space. In contrast, for anomalous videos, it maximizes the distance between feature clusters, enabling the features of anomalous videos to be distinctly separated from those of normal videos in the feature space. 
	Wu~\etal~\cite{wu2024vadclip} introduce VadCLIP, which directly utilizes the frozen CLIP model without any pre-training or fine-tuning. In contrast to existing methods that simply input extracted features into a weakly supervised classifier for frame-level binary classification, VadCLIP fully exploits the fine-grained associations between vision and language enabled by CLIP, incorporating a dual-branch architecture.
	Chang~\etal~\cite{chang2021contrastive} develop a lightweight anomaly detection model that uses normal videos to train a classifier with strong discriminative ability, enhancing the selectivity for anomalous segments and filtering out normal ones. Shi~\etal~\cite{shi2023abnormal} propose a novel framework for WS-VAD, termed ARMS, featuring an abnormal ratio-based MIL loss (AR-MIL) and a multi-phase self-training paradigm. Unlike the aforementioned methods, this paper proposes a novel text-guided multimodal video anomaly detection framework, which introduces the text modality under a weakly supervised setting to provide robust detection guidance.

	\subsection{Weakly Supervised Multimodal Video Anomaly Detection}
	Weakly supervised multimodal video anomaly detection (WS-MVAD)~\cite{wei2022msaf,chen2023tevad,yuan2024surveillance}, also known in some works as weakly supervised multimodal violence detection (WS-MVD)~\cite{wu2020not,wu2022weakly,yu2022modality,pang2022audiovisual,wei2022look,sun2024multi}, refers to the use of additional modalities beyond RGB or optical flow, such as audio or text modality, to detect anomalies under the weak supervision setting. Compared to methods that utilize only RGB or optical flow, designing a framework that incorporates multiple modalities, such as audio and text features, can significantly enhance the performance of anomaly detection~\cite{yu2022modality,pang2022audiovisual}. Numerous efforts have been devoted to integrating several modalities, resulting in the development of various fusion techniques. These include methods like concatenation or addition~\cite{wu2020not,wu2022weakly,chen2023tevad}, bilinear pooling~\cite{pang2021violence}, co-attention~\cite{wei2022look}, and cross-attention~\cite{liu2022decouple,yu2022modality}. For example,
	Pang~\etal~\cite{pang2021violence} propose a neural network, which consists of an attention module, a fusion module, and a mutual learning module, to investigate fusion methods that combine visual and audio features for violence detection. Experimental results demonstrate the effectiveness of each module, and their combination significantly enhances the model's overall performance on the XD-Violence dataset, achieving 81.69\% AP performance.
	Yu~\etal~\cite{yu2022modality} use cross-modality attention to fuse the two modalities, producing the fused features. Chen~\etal~\cite{chen2023tevad} propose a text-empowered video anomaly detection (TEVAD) framework, which introduces text features into the video anomaly detection. However, this method directly extracts the text features by the vanilla text feature extractor and employs the concatenation and addition methods to fuse the text and visual features, not yield particularly satisfactory results.
	Recently, Yuan~\etal~\cite{yuan2024towards} meticulously annotated the real-world surveillance dataset UCF-Crime with detailed event content and timing. Through experiments on this dataset, they demonstrate that leveraging it for multimodal surveillance learning enhances anomaly detection performance. 
	Furthermore, they select TEVAD~\cite{chen2023tevad} as the baseline for multimodal video anomaly detection~\cite{yuan2024surveillance}, which integrates multimodal features, such as visual, temporal, and caption features. Additionally, they propose a novel anomaly detection framework that incorporates captions, building upon TEVAD, which improves its anomaly detection performance.
	Tan~\etal~\cite{tan2024overlooked} investigate the effectiveness of whole-video classification supervision using BERT or LSTM. By utilizing BERT or LSTM, CNN features from all video snippets are aggregated into a single feature, which can then be used for whole-video classification. This straightforward yet effective supervision, when combined with the MIL and RTFM frameworks, delivers outstanding performance on video anomaly datasets, achieving an 86.71\% AUC on UCF-Crime by integrating RGB and flow modalities.
	Shi~\etal~\cite{shi2024caption} use video memory to represent both video features (non-semantic) and caption information (semantic). Their method generates and updates memory during training, predicting anomaly scores based on memory similarities between the input video and stored memories, and captions serve as descriptions of representative anomaly actions.
	Na~\etal~\cite{na2024leveraging} extend the dual-modality TEVAD~\cite{chen2023tevad} to a multi-modal framework integrating visual, audio, and textual data. They also refine the multi-scale temporal network (MTN) to improve feature extraction across multiple temporal scales in video snippets.
	Xiao~\etal~\cite{xiao2023scoreformer} introduce a score fusion-based Transformer framework called Scoreformer. Specifically, optical flow, RGB, and audio features are first processed through independent self-attention Transformer blocks. Then, the optical flow and RGB features are passed through cross-modal Transformer blocks, after which they are fused with the audio features using a score fusion block. This approach prevents the noise interference that could arise from directly fusing audio and visual features. This method combines RGB, flow, and audio modalities, achieving an AP performance of 84.54\% on the XD-Violence dataset. The majority of these efforts concentrate on audio-visual learning, leaving the efficient integration of additional modalities largely unexplored. In contrast, we propose an MSTA mechanism to fine-tune the text feature extractor, obtaining high-quality text features. Besides, we introduce a multi-scale bottleneck Transformer (MSBT) that accommodates a variable number of modalities for the WS-MVAD task, achieving state-of-the-art performance by utilizing RGB, optical flow, audio, and text modalities.

	\subsection{Multimodal Fusion}
	Cross-attention based multimodal fusion techniques have been applied in various vision tasks, such as fine-grained recognition~\cite{wei2024mecom} and temporal action localization~\cite{hong2021cross}. For example, Wei~\etal~\cite{wei2024mecom} propose a meta-completion method called MECOM for the task of fine-grained recognition, which integrates multimodal fusion and explicitly completes missing modalities through their techniques of cross-modal attention and decoupling reconstruction. To enhance fine-grained recognition accuracy, they introduce an additional partial stream, complementing the main stream of MECOM, along with a part-level feature selection mechanism that targets fine-grained object parts.
	Hong~\etal~\cite{hong2021cross} propose a cross-modal consensus network for temporal action localization. The approach introduces two identical cross-modal consensus modules, each incorporating a cross-modal attention mechanism. This mechanism filters out task-irrelevant redundant information by leveraging global information from the primary modality and cross-modal local information from the auxiliary modality.
	
	Due to the inherent characteristics of self-attention and its variants, Transformers can function in a modality-agnostic manner and are scalable to multiple modalities~\cite{xu2023multimodal}. Consequently, multimodal Transformers have been widely adopted to integrate diverse modalities such as video, audio, and text across various applications, including emotion recognition~\cite{li2023decoupled}, speech recognition~\cite{song2022multimodal,tzirakis2021end}, video retrieval~\cite{dzabraev2021mdmmt,gabeur2020multi,Shvetsova2022Transformer}, visual question answering~\cite{Urooj2020Transformer}, and autonomous driving~\cite{Prakash2021Transformer,lin2022multimodal}, among others. For instance, Li~\etal~\cite{li2023decoupled} propose a decoupled multimodal distillation (DMD) approach to enable flexible cross-modal knowledge distillation and enhance discriminative features. Each modality is decoupled into modality-irrelevant and modality-exclusive spaces through self-regression. Besides, this method uses multimodal Transformers to bridge the distribution gap. DMD uses a graph distillation unit (GD-Unit) for each decoupled part, where a dynamic graph represents modalities and their knowledge distillation. This approach allows automatic learning of distillation weights, supporting diverse knowledge transfer patterns.
	
	However, the efficiency of multimodal Transformers becomes a challenge as the number of modalities increases. To reduce information redundancy, Nagrani~\etal~\cite{nagrani2021attention} proposed a multimodal bottleneck Transformer (MBT) that uses a small set of bottleneck tokens for modality interaction, enhancing fusion performance. Inspired by MBT~\cite{nagrani2021attention}, we propose a multi-scale bottleneck Transformer (MSBT) and a bottleneck token-based weighting scheme for multimodal fusion. Unlike MBT~\cite{nagrani2021attention}, which employs a fixed number of bottleneck tokens used solely as intermediaries for modality interaction, our MSBT utilizes a decreasing number of tokens for gradual condensation and leverages the learned tokens to weight fused features, thereby achieving more effective fusion performance.
	
	\begin{figure*}[t]
		\centering
		\includegraphics[width=0.97\textwidth]{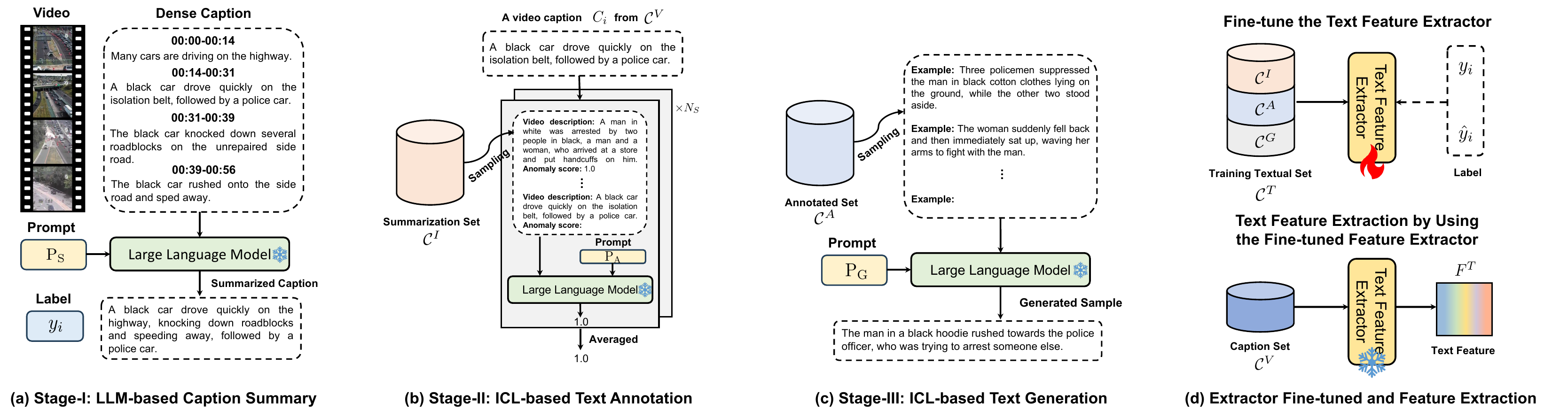} 
		\caption{Illustration of (a$\sim$c) the proposed multi-stage text augmentation (MSTA) for (d) extractor fine-tuning and feature extraction. (a) Stage-I: We use a large language model (LLM) to summarize all captions in the videos, obtaining labeled text samples. (b) Stage-II: based on the summarized captions, we utilize ICL to generate pseudo-labels for each caption within the video. (c) Stage-III: We employ the labeled samples from the previous two stages, using ICL to generate new anomalous samples. (d) We fine-tune the feature extractor using both the original and generated samples to obtain high-quality text representation features.}	
		\label{fig:multi-stage}
		\vspace{-10pt}
	\end{figure*}
	
	\subsection{Text Augmentation}
	Text augmentation involves techniques for perturbing the linguistic space without changing class labels, aimed at enhancing the robustness and generalizability of downstream models in the natural language processing (NLP) tasks~\cite{moller2023prompt,zhang2023recommendation,peng2023instruction}.  For instance, M{\o}ller~\etal~\cite{moller2023prompt} provide an example along with its corresponding label and guide the large language models (LLMs) to generate similar examples with the same label, leading to improved downstream performance in few-shot classification tasks. Zhang~\etal~\cite{zhang2023recommendation} create a comprehensive dataset of user-personalized instructions and fine-tune the LLMs with them, resulting in more accurate recommendations that outperform competitive benchmarks. Peng~\etal~\cite{peng2023instruction} use GPT-4 to create an instruction-following dataset and feedback data for tuning the model. The instruction-tuned LLaMA models from this dataset achieved performance comparable to GPT-4. Although text augmentation has been widely applied in the aforementioned tasks, its potential in the task of VAD remains unexplored. In this paper, we propose a multi-stage text augmentation method that generates a large number of anomalous samples. This approach aims to address the imbalance between normal and anomalous samples in the VAD task.
	
	\subsection{In-context Learning}
	As large language models (LLMs) advance, in-context learning (ICL) has emerged as a new NLP paradigm, where LLMs make predictions based on context enriched with a few examples. Exploring LLMs through ICL has become a prominent trend~\cite{dong2022survey}. ICL bypasses parameter updates and instead relies on pre-trained language models to generate predictions directly. The model is designed to identify and learn patterns within the given demonstrations, enabling it to make accurate predictions accordingly. Min~\etal~\cite{min2022rethinking} demonstrate that ground truth demonstrations are not essential, as randomly substituting labels in the demonstrations has little impact on performance across a variety of classification tasks. Li~\etal~\cite{li2023few}  propose KB-BINDER, which enables few-shot in-context learning over knowledge-based question-answering (KBQA) tasks, addressing questions over diverse KBQA datasets with a unified training-free framework. In this work, we propose a multi-stage text augmentation (MSTA) method based on ICL, aimed at generating pseudo-labels for text and producing anomalous text samples.
	
	\section{Multi-stage Text Augmentation for Text Encoder Fine-tuning}
	Pre-trained large language models are not well-suited for anomaly detection because they are trained on general datasets, focusing on common patterns, and lack sensitivity to rare abnormal events, \ie they miss subtle, domain-specific anomalies. As our experimental results show in Section~\ref{sec:ab_MSTA}, we observed that directly applying the text features extracted by the vanilla feature extractor to anomaly detection did not yield satisfactory performance. These reasons and observations encourage us to fine-tune a text feature extractor for the specific task of video anomaly detection. However, the scarcity of anomalous events results in a predominance of normal text samples within the video data. This imbalance in the dataset introduces potential bias during the fine-tuning process, ultimately hindering the ability to obtain an effective feature extractor. Fortunately, text samples are relatively easy to generate. We can achieve a balanced sample size by producing anomalous text samples, which facilitates the fine-tuning of feature extractors. Therefore, we propose a multi-stage text augmentation (MSTA) for text samples augmentation, as shown in Fig.~\ref{fig:multi-stage}. This involves using in-context learning to generate a good deal of text descriptions of anomalous events in the dataset, after which the generated anomalous samples are used to fine-tune the pre-trained text feature extractor to extract more distinguishable features. 
	
	In the \text{Stage-I} of MSTA, a summary of all captions in the video allows us to obtain labeled text samples. In the  \text{Stage-II}, based on the summarized captions, we utilize ICL to produce the anomaly scores as pseudo-labels for each caption within the video. The  \text{Stage-III} employs the labeled samples from the previous two stages, using ICL to generate new anomalous samples, thereby achieving a balance between the quantities of normal and anomalous samples. The details of each stage in MSTA are as follows.
	
	\subsection{Stage-I: LLM-based Caption Summary}
	Unlike the RGB or audio information explicitly present in videos, for the text information, we need to leverage a pre-trained video captioning model to generate dense captions for the videos, \ie the caption set for the $i$-th video is denoted as $\mathcal{C}_i=\{C_{i,1}, C_{i,2}, ..., C_{i,N}\}$. In the task of weakly supervised video anomaly detection, since each video has a video-level annotation, it is unable to determine whether $C_{i,j}$ specifically describes an anomalous event. Fortunately, we can leverage the powerful text summarization capabilities of LLM to condense all the captions of a video into a concise sentence that summarizes the entire video. Given that we have video-level labels, the summarized sentence inherits a definitive label, \ie when $y_i=1$, the summarized sentence describes an anomalous event, and if $y_i=0$, it does not. Therefore, the summarized caption for the $i$-th video is as follows:
	\begin{equation} 
		C^I_i = \phi_{LLM}(\text{P}_\text{S}\diamond C_{i,1}\diamond C_{i,2}\diamond ... \diamond C_{i,N}),
	\end{equation}
	where $\diamond$ operator denotes that concatenates the texts, $\text{P}_\text{S}$ is the prompt used for summarization by the LLM, which takes the form ``\textit{Please summarize the following sentences into a single sentence of no more than 30 words. Please just output the summarized sentence without additional details or introductions, just one sentence.}''. The summaried texts of all videos form the summarization set $\mathcal{C}^I$, \ie $\mathcal{C}^I=\{C^I_i\}^{N_D}_{i=1}$.
	
	\subsection{Stage-II: ICL-based Text Annotation.}
	From the \text{Stage-I} of processing, we obtain labeled text samples, with these samples being equivalent in number to the training videos. In order to use a more diverse set of samples as a reference during the \text{Stage-III} of text generation, we aim to increase the number of labeled samples. 
	In detail, for the set of captions $\mathcal{C}_i$ in each video, each caption is unlabeled, meaning it is impossible to determine whether it describes a normal or anomalous event. The objective of Stage II is to assign pseudo-labels to these captions, thereby increasing the number of labeled samples.
	To this end, we propose a method based on in-context learning that employs multiple sampling and scoring iterations, assigning an anomaly score as a pseudo-label to each caption within a video. This anomaly score functions as a pseudo-label for the captions. To be specific, in the process of labeling each caption in a video, we perform $N_S$ samplings of the set $\mathcal{C}^I$. For each sampling, $N_R$ context samples are randomly selected from $\mathcal{C}^I$ and then formatted to serve as the demonstration context. Thus, each sample is formatted as follows:
	\begin{equation} 
		C^{II}_j = \text{P}_\text{VD}\diamond C^I_i\diamond\text{P}_\text{AS}\diamond y_i, i\in\mathcal{I}, j\in[1,N_R],
	\end{equation}
	where the forms of $ \text{P}_\text{VD}$ and $\text{P}_\text{AS}$ are ``\textit{Video description:}'' and ``\textit{Anomaly score:}'', respectively, the $y_i$ is the video-level label and $\mathcal{I}$ denotes the indices set of the selected samples. Therefore, the complete context for $i$-th sampling is:
	\begin{equation} 
		\hat{C}^{II}_i =  C^{II}_1\diamond C^{II}_2\diamond ... \diamond C^{II}_{N_R}.
	\end{equation}
	
	By utilizing the concatenated $\hat{C}^{II}_i$ as the context, we can employ in-context learning with a LLM to generate an anomaly score for each caption $C_i\in\mathcal{C}^{V}$=$\{C_j\in\mathcal{C}_k|k\in[1,N_D]\}$. The pseudo-label for each caption is then determined by averaging the anomaly scores obtained from multiple samplings:
	\begin{equation} 
		\hat{y}_i =  \frac{1}{N_S}\sum\limits_{k=1}^{N_S} \phi_{LLM}(\text{P}_\text{A}\diamond \hat{C}^{II}_k\diamond C_i\diamond \text{P}_\text{AS}),
	\end{equation}
	where $\hat{y}_i$ serves as the pseudo-label of the $i$-th caption, $\text{P}_\text{A}$ is formed as ``\textit{Each element in the following list contains a description of a video and the corresponding anomaly score. The anomaly score indicates the probability of an anomalous event occurring in the video. Just complete the last space of the correct anomaly score.}'', and $\hat{C}^{II}_k$ denotes the demonstration context of the $k$-th sampling.
	
	\subsection{Stage-III: ICL-based Text Generation.}
	To generate more text descriptions of anomalous events, we also employ an in-context learning based approach, leveraging the large number of labeled anomaly samples produced in \text{Stage-II} as context samples. Through \text{Stage-II}, we obtain a set $\mathcal{C}^A=\{C_j\in\mathcal{C}^V|\hat{y}_j\textgreater\delta \}$ composed of anomaly event descriptions, where the $\delta$ is a hyperparameter. To produce a new anomaly description, we randomly select $N_R$ samples from $\mathcal{C}^A$ to serve as the demonstration context, and each sample is then formatted as follows:
	\begin{equation} 
		C^{III}_j = \text{P}_\text{EX}\diamond C_i, C_i\in\mathcal{C}^A,
	\end{equation}
	where the $\text{P}_\text{EX}$ is formed as ``\textit{Example:}''. Thus, the complete context for the $i$-th generated sample is:
	\begin{equation} 
		\hat{C}^{III}_i =  C^{III}_1\diamond C^{III}_2\diamond ... \diamond C^{III}_{N_R}.
	\end{equation}
	
	Drawing upon the complete context, the description of the $i$-th newly generated anomaly event by the LLM can be expressed as follows:
	\begin{equation} 
		\tilde{C}^{III}_i = \phi_{LLM}(\text{P}_\text{G}\diamond \hat{C}^{III}_i\diamond\text{P}_\text{EX}),
	\end{equation}
	where $\text{P}_\text{G}$ is the generation prompt formed as ``\textit{Each element in the following list contains a description of a video snippet whose category is abnormal events. Just generate one sentence in the same category as the above sentences.}''. The pseudo-label $\hat{y}_i$ for the generated text in $\tilde{C}^{III}_i$ is set to $1$. Following $N_G$ iterations of generation, we derive the resulting set of generated samples, denoted as $\mathcal{C}^{G}=\{\tilde{C}^{III}_i\}^{N_G}_{i=1}$.
	
	\section{Weakly Supervised Multimodal Video Anomaly Detection}
	
	\subsection{Problem Formulation}
	In the task of WS-MVAD, there is a training set $\mathcal{V}^T$ including $N_D$ videos with varied durations and a test set $\mathcal{V}^E$. Each training video is annotated by a binary label $y\in\{0,1\}$, indicating whether it is normal or anomalous, and the labels for all training videos together form the label set $\mathcal{Y}$. The goal of this task is to train a model $f(\cdot)$ that takes various modalities of a video as input, such as RGB, audio, optical flow, or text, and the model is required to generate a frame-level anomaly score to predict whether each video frame is classified as normal or abnormal.

	\subsection{Feature Extraction}
	
	\subsubsection{Video Feature Extraction}
	In multimodal video anomaly detection methods~\cite{wu2020not, yu2022modality, sun2024multi}, pre-trained models are typically employed to extract high-level features from various segments of a video. Following prior works~\cite{yu2022modality, pang2022audiovisual,liu2022decouple,wu2022weakly}, we extract RGB and flow features utilizing the I3D~\cite{carreira2017quo} model pre-trained on the Kinetics-400 dataset and extract audio features employing the VGGish~\cite{gemmeke2017audio} pre-trained on the YouTube dataset. Specifically, each video is divided into several segments, with each segment consisting of a consecutive and non-overlapping set of $N_C$ frames (\eg $N_C$=16). These $N_C$ frames with corresponding audio information are then fed into pre-trained models to produce RGB, flow, and audio embedding features, denoted as $F^R\in\mathbb{R}^{D_R}$, $F^F\in\mathbb{R}^{D_F}$, and $F^A\in\mathbb{R}^{D_A}$, respectively, where $D_R$, $D_F$, and $D_A$ represent the dimensions of the respective features.
	
	\subsubsection{Text Feature Extraction}
	
	\begin{figure}[t]
		\centering
		\includegraphics[width=0.48\textwidth]{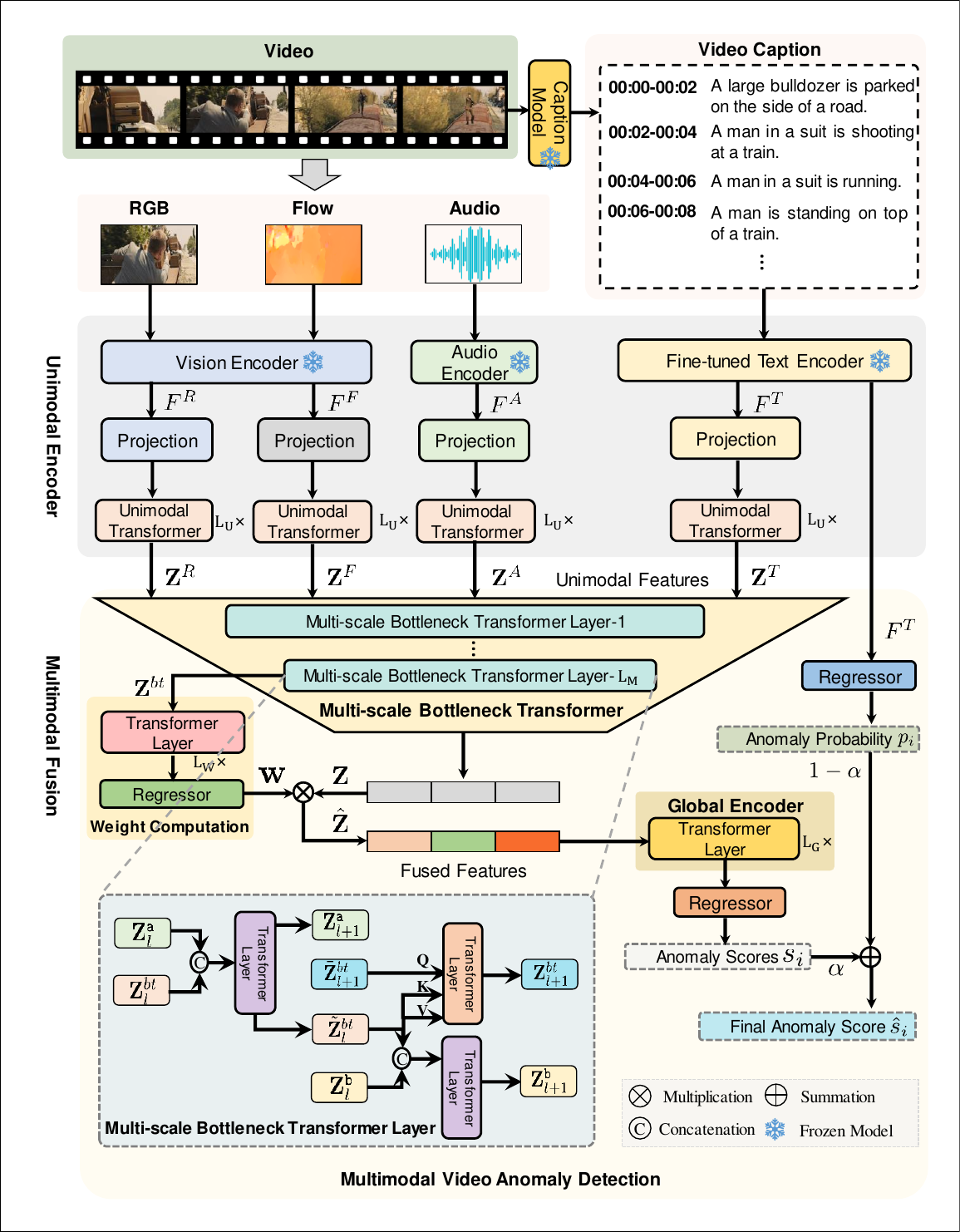} 
		\caption{An overview of the proposed framework. It includes three unimodal encoders, a multi-scale bottleneck Transformer, and a global encoder for multimodal feature generation. Each unimodal encoder consists of a modality-specific feature extractor and a linear projection layer for tokenization and a modality-shared Transformer for context aggregation within one modality. The multi-scale bottleneck Transformer (MSBT) fuses any pair of modalities and a sub-module to weight concatenated fused features. The global encoder, implemented by a Transformer, aggregates context overall snippets. Finally, the final anomaly score is constructed by combining the anomaly scores from the fused features and the anomaly probabilities from the text. }	
		\label{fig:framework}
		\vspace{-10pt}
	\end{figure}
	
	Through \text{Stage-III} of MSTA, we can generate an equal number of anomalous event descriptions to match the number of normal event descriptions. This balanced set of samples allows us to fine-tune the text feature extractor $\phi_{TFE}(\cdot)$ well, enhancing its ability to distinguish between normal and anomalous text features. To fine-tune the feature extractor, we append a regressor $\phi_{PR}(\cdot)$ to enable it to produce anomaly probability:
	\begin{equation}  \label{eq:text_anomaly_score}
		p_i =\phi_{PR}( \phi_{TFE}(C_i)),
	\end{equation}
	where $\phi_{TFE}(\cdot):\mathcal{T}\rightarrow\mathbb{R}^{D_T}$  extracts the input text into the feature of dimensionality $D_T$, $\phi_{PR}(\cdot):\mathbb{R}^{D_T}\rightarrow\mathbb{R}$ transforms the text feature into an anomaly score, implemented using a two-layer multi-layer perceptron (MLP) followed by a sigmoid activation function, and $C_i\in\mathcal{C}^{T}=\mathcal{C}^{I}\cup\mathcal{C}^{V}\cup\mathcal{C}^{G}$ is a text sample from the combined training text set $\mathcal{C}^{T}$, which merges $\mathcal{C}^{V}$, $\mathcal{C}^{I}$, and $\mathcal{C}^{G}$.
	
	Utilizing the labels $\hat{\mathcal{Y}}$ and text samples, we fine-tune the $\phi_{TFE}(\cdot)$ by minimizing the binary cross-entropy (BCE) loss:
	\begin{equation}  \label{eq:loss_TFE}
		\mathcal{L}_{TFE}=-\hat{y}_ilog(p_i)-(1-\hat{y}_i)log(1-p_i),
	\end{equation}
	where $\hat{y}_i\in\hat{\mathcal{Y}}=\{y_i\}^{N_D}_{i=1}\cup\{\hat{y}_i\}^{|\mathcal{C}^{V}|+|\mathcal{C}^{G}|}_{i=1}$ is the pseudo-label of $i$-th text sample.
	
	Finally, we use the fine-tuned $\phi_{TFE}(\cdot)$ to extract the text feature $F^T\in\mathbb{R}^{D_T}$, as shown in Fig.~\ref{fig:multi-stage}(d).

	\subsection{MSBT-based Multimodal Video Anomaly Detection}
	The proposed text-guided multimodal video anomaly detection (TG-MVAD) framework is shown in Fig.~\ref{fig:framework}. Each video is divided into $N_T$ non-overlapping snippets. The video caption is obtained by a pre-trained caption model. The RGB, flow, audio, or text snippets go through an unimodal encoder composed of a pre-trained encoder and a linear projection layer for tokenization, and an unimodal Transformer for intra-modal context aggregation, by which modality-specific features are generated. Then, the modality-specific features are fed into a fusion module that employs a multi-scale bottleneck Transformer to fuse features pairwise for all modality pairs and further concatenates and weights the fused features based on the learned bottleneck tokens. The fused features finally pass through a global encoder and a regressor to predict anomaly scores. Besides, the text modality is processed through a regressor to yield the anomaly probabilities. The final anomaly score is constructed by combining the anomaly scores from the fused features and the anomaly probabilities from the text.

	\subsubsection{Unimodal Encoder}
	We construct three unimodal encoders to produce modality-specific features from input snippets. Each input snippet, depending on its modality, is tokenized using a pre-trained feature extraction backbone (e.g., the I3D model~\cite{carreira2017quo} for RGB or optical flow, the VGGish model~\cite{gemmeke2017audio} for audio, and the BERT~\cite{devlin2018bert} for text), followed by a linear projection layer that outputs an embedded feature vector $\mathbf{z} \in\mathbb{R}^{1\times D_E}$, where ${D_E}$ is the feature dimension unified for all modalities. Subsequently, we use a vanilla Transformer~\cite{vaswani2017attention} to aggregate contextual information within each modality. Specifically, the embedded feature is input into an unimodal Transformer with $L_U$ layers, where each layer comprises Multi-head Self-Attention (MSA) and FeedForward Network (FFN) blocks. The computation for the $l$-th Transformer layer, $\mathbf{z}_{l+1} = \text{Transformer}(\mathbf{z}_{l})$, is defined as:
	\begin{equation} \label{eq:MSA_FFN}
		\begin{aligned}
			\hat{\mathbf{z}}_{l} &= \text{MSA}(\text{LN}(\mathbf{z}_{l})) + \mathbf{z}_{l}, \\
			\mathbf{z}_{l+1} &= \text{FFN}(\text{LN}(\hat{\mathbf{z}}_{l})) + \hat{\mathbf{z}}_{l},
		\end{aligned}
	\end{equation}
	where LN denotes layer normalization and FFN is implemented through a two-layer Multilayer Perceptron (MLP). The self-attention mechanism for each head within MSA calculates dot-product attention as follows:
	\begin{equation}
		\begin{aligned}
			\text{MSA}(\mathbf{x}) &= \text{Attention}(\mathbf{x}\mathbf{W}^{Q}, \mathbf{x}\mathbf{W}^{K}, \mathbf{x}\mathbf{W}^{V}) \\
			&= \text{SoftMax}\left(\frac{\mathbf{x}\mathbf{W}^{Q}(\mathbf{x}\mathbf{W}^{K})^{T}}{\sqrt{d}}\right) \mathbf{x}\mathbf{W}^{V},
		\end{aligned}
	\end{equation}
	where $d = D_E$, and $\mathbf{W}^{Q}$, $\mathbf{W}^{K}$, $\mathbf{W}^{V}\in\mathbb{R}^{D_E\times D_H}$ are learnable matrices such that $D_H = D_E/N_H$, with $N_H$ being the number of attention heads. In the unimodal encoders, the feature extractor and linear projection layer are unique to each modality, whereas the unimodal Transformer is shared across all modalities.

	\subsubsection{Multimodal Fusion}
	The multimodal fusion is performed at the video level. Let the modality-specific features of a video, produced by the unimodal encoders, be represented as $\mathbf{Z}^R$, $\mathbf{Z}^F$, $\mathbf{Z}^A$, and $\mathbf{Z}^T$ for RGB, flow, audio, and text modalities, respectively. Specifically, $\mathbf{Z}^{\mathtt{a}} = [\mathbf{z}^{\mathtt{a}}_{{L_{U+1}}(1)}, ..., \mathbf{z}^{\mathtt{a}}_{{L_{U+1}}(N_T)}] \in \mathbb{R}^{N_T\times D_E}$, where $\mathtt{a} \in \mathcal{M} = \{T, R, F, A\}$ represents a modality, $N_T$ is the number of snippets, and $\mathbf{z}^{\mathtt{a}}_{{L_{U+1}}}$ is the output from the unimodal Transformer. We then conduct multimodal fusion using a multi-scale bottleneck Transformer (MSBT) and a bottleneck token-based weighting scheme as outlined below.
	
	\noindent\textbf{Multi-scale Bottleneck Transformer.} 
	Bottleneck Transformers~\cite{nagrani2021attention,Yan2022Transformer} have shown both efficiency and effectiveness in multimodal fusion by utilizing a small number of bottleneck tokens to condense the information of a modality, thus significantly reducing information redundancy. Drawing inspiration from these works, we propose a multi-scale bottleneck Transformer (MSBT) with $L_M$ layers to fuse the features of each pair of modalities. Our multi-scale strategy employs a progressively decreasing number of bottleneck tokens at different layers, allowing for gradual information condensation. This approach contrasts with the fixed number of tokens at all layers used in~\cite{nagrani2021attention,Yan2022Transformer}, leading to more effective fusion performance.
	
	More specifically, given modality-specific features $\mathbf{Z}^{\mathtt{a}}$ and $\mathbf{Z}^{\mathtt{b}}$ for two different modalities $\mathtt{a}$ and $\mathtt{b}$, respectively, where $\mathtt{a}, \mathtt{b} \in \mathcal{M}$, the MSBT fuses the information from modality $\mathtt{a}$ into modality $\mathtt{b}$ at layer $l$ by
	\begin{equation} \label{eq:MSBT_a_bt}
		[\mathbf{Z}^{\mathtt{a}}_{l+1}, \tilde{\mathbf{Z}}^{bt}_l] = \text{Transformer}([\mathbf{Z}^{\mathtt{a}}_{l}||\mathbf{Z}^{bt}_l]),
	\end{equation}
	\begin{equation} \label{eq:MSBT_b_bt}
		[\mathbf{Z}^{\mathtt{b}}_{l+1}, \hat{\mathbf{Z}}^{bt}_{l}] = \text{Transformer}([\mathbf{Z}^{\mathtt{b}}_{l}|| \tilde{\mathbf{Z}}^{bt}_l]), 
	\end{equation}
	where $[\cdot||\cdot]$ denotes the feature concatenation operation. The term $\mathbf{Z}^{bt}_l\in\mathbb{R}^{N^{bt}_{l}\times D_E}$ ($N^{bt}_{l} \ll T$) represents the bottleneck tokens input to layer $l$, which carry the previously condensed information of modality $\mathtt{a}$ and are learned from the preceding layer ($\mathbf{Z}^{bt}_1$ is randomly initialized). At the current layer, these tokens serve as intermediaries for transmitting information from modality $\mathtt{a}$ to modality $\mathtt{b}$. First, they are refined by re-aggregating information from modality $\mathtt{a}$ via Eq.~(\ref{eq:MSBT_a_bt}). Then, they transmit this information to modality $\mathtt{b}$ using Eq.~(\ref{eq:MSBT_b_bt}), resulting in the fusion output $\mathbf{Z}^{\mathtt{b}}_{l+1}$.
	
	Additionally, in our proposed MSBT, the bottleneck tokens also serve to pass the condensed information of modality $\mathtt{a}$ from the current layer to the next layer. To facilitate this, we initialize a new set of bottleneck tokens $\bar{\mathbf{Z}}^{bt}_{l+1}\in\mathbb{R}^{N^{bt}_{l+1}\times D_E}$ at layer $l$. The number of tokens is halved, i.e., $N^{bt}_{l+1} = \lfloor N^{bt}_{l}/2 \rfloor$, to store more condensed information. The condensed information from layer $l$ is then passed to layer $l+1$ by 
	\begin{equation} \label{eq:MSBT_CT}
		\mathbf{Z}^{bt}_{l+1} = \text{Cross-Transformer}(\bar{\mathbf{Z}}^{bt}_{l+1}, \tilde{\mathbf{Z}}^{bt}_l).
	\end{equation}
	Here, $\text{Cross-Transformer}(\mathbf{x}, \mathbf{y})$ is a cross-attention based Transformer that shares the same architecture as $\text{Transformer}(\mathbf{x})$ but computes the following attention:
	\begin{equation}
		\text{Attention}(\mathbf{x}\mathbf{W}^Q, \mathbf{y}\mathbf{W}^K, \mathbf{y}\mathbf{W}^V).
	\end{equation}
	
	When fusing the modality-specific feature $\mathbf{Z}^{\mathtt{a}}$ into $\mathbf{Z}^{\mathtt{b}}$, we take $\mathbf{Z}^{\mathtt{b}}_{L_M+1}$ as the fusion result and denote it as $\mathbf{Z}^{\mathtt{a}\mathtt{b}}$. This process essentially transmits the condensed information from modality $\mathtt{a}$ to modality $\mathtt{b}$, indicating that the fusion of two modalities is asymmetric. Consequently, after fusing all pairs of modalities, we obtain $N_F=A^2_{|\mathcal{M}|}$ fused features, \eg $\mathbf{Z}^{TR}$, $\mathbf{Z}^{RF}$, and $\mathbf{Z}^{FA}$, where $A^n_{m}$ represents the number of permutations when selecting $n$ ones from $m$ elements, where $|\mathcal{M}|$ denotes the size of the set $\mathcal{M}$. 
	
	\noindent\textbf{Bottleneck Token-based Weighting.} 
	Through the MSBT module, we obtain several feature pairs resulting from the fusion of two different modalities, such as $\mathbf{Z}^{TR}$, $\mathbf{Z}^{RF}$, and so on. These feature pairs are then merged to form the final multimodal fused feature.
	A straightforward approach to obtaining the fused feature from all modalities is to concatenate the features that have been fused from all pairwise combinations. That is,
	\begin{equation} \label{eq:fusion_prior}
		\begin{aligned}
			\mathbf{Z} =[&\mathbf{Z}^{TR}||\mathbf{Z}^{RT}||\mathbf{Z}^{TF}||\mathbf{Z}^{FT}||\mathbf{Z}^{TA}||\mathbf{Z}^{AT}||\\
			&\mathbf{Z}^{RF}||\mathbf{Z}^{FR}|| \mathbf{Z}^{RA}||\mathbf{Z}^{AR}||\mathbf{Z}^{FA}||\mathbf{Z}^{AF}],
		\end{aligned}
	\end{equation}
	where $\mathbf{Z} \in \mathbb{R}^{N_T\times (N_F\cdot D_E)}$, 
	$N_T$, $N_F$, and $D_E$ represent the number of segments in the current video, the number of modality pair features after fusion through the MSBT module, and the dimensionality of the features, respectively. However, we recognize that the bottleneck token learned at the final layer, $\tilde{\mathbf{Z}}^{bt}_{L_M}$, can be utilized to measure the amount of information transferred from one modality to another. Consequently, we propose leveraging these learned bottleneck tokens to weight the features fused from all pairs of modalities. The weights $\mathbf{w} = [w_1, ..., w_{N_F}] \in \mathbb{R}^{1 \times N_F}$ are derived by inputting the tokens into an $L_W$-layer Transformer followed by a regressor. That is,
	\begin{equation} \label{eq:MSBT_W}
		\mathbf{w} =\phi_{WR}(\text{Transformer}_{(L_W\times)}(\mathbf{Z}^{bt})),
	\end{equation}
	where $\mathbf{Z}^{bt}\in\mathbb{R}^{N_T\times N_F\times D_E}$ is a stack of the final bottleneck tokens acquired from all pairwise fusions, $\text{Transformer}_{(L_W\times)}(\cdot)$ represents an $L_W$-layer Transformer, and $\phi_{WR}(\cdot)$ is a regressor implemented using a three-layer MLP. The weighted feature is then obtained by
	\begin{equation}
		\begin{aligned}
			\hat{\mathbf{Z}}=&[w_1\mathbf{Z}^{TR}||w_2\mathbf{Z}^{RT}||w_3\mathbf{Z}^{TF}||w_4\mathbf{Z}^{FT}||w_5\mathbf{Z}^{TA}||w_6\mathbf{Z}^{AT}||\\
			&w_7\mathbf{Z}^{RF}||w_8\mathbf{Z}^{FR}||w_9\mathbf{Z}^{RA}||w_{10}\mathbf{Z}^{AR}||w_{11}\mathbf{Z}^{FA}||w_{12}\mathbf{Z}^{AF}],
		\end{aligned}
	\end{equation}
	where $\hat{\mathbf{Z}} \in \mathbb{R}^{N_T\times (N_F\cdot D_E)}$ represents the fused features of all snippets in a video.
	
	\noindent\textbf{Anomaly Score.}
	Once the fused features from all snippets of a single video are generated, we utilize a global encoder, which is implemented using a $L_G$-layer Transformer, to capture the global context across all snippets in a video, as seen in~\cite{xu2023multimodal,Zhan2021}. We then apply a regressor to generate fused anomaly scores, \ie
	\begin{equation} \label{eq:fused_anomaly_score}
		\mathbf{s}=\phi_{SR}(\text{Transformer}_{(L_G\times)}(\hat{\mathbf{Z}})).
	\end{equation}
	Here, $\mathbf{s}\in\mathbb{R}^{T}$ represents the anomaly scores for all snippets in a video, and $\phi_{SR}(\cdot)$ is the regressor implemented using a three-layer MLP.

	\subsection{Training and Inference}
	
	\noindent\textbf{Network Training.}
	The training process is divided into two phases: 1) fine-tuning the text feature extractor and 2) training the MSBT module. We fine-tune the text feature extractor $\phi_{TFE}$ by minimizing $\mathcal{L}_{TFE}$ as shown in Eq.~(\ref{eq:loss_TFE}), and we utilize top-$K$ MIL loss~\cite{tian2021weakly} to train the MSBT, enabling the model to classify the normal and abnormal events. We fine-tune the text feature extractor $\phi_{TFE}$ by minimizing $\mathcal{L}_{TFE}$ as shown in Eq.~(\ref{eq:loss_TFE}), and we utilize top-$K$ MIL loss~\cite{tian2021weakly} to train the MSBT, enabling the model to classify the normal and abnormal events. To be specific, we average the top-$K$ anomaly scores as follows,
	\begin{equation}
		\bar{s} =\frac{1}{K}\sum\nolimits_{s_i\in\mathcal{S}_K(\mathbf{s})}s_i,
	\end{equation}
	where $\mathcal{S}_K(\mathbf{s})$ indicates the set of top-$K$ scores in $\mathbf{s}$.  Similar to Eq.~\ref{eq:loss_TFE}, the tok-$K$ MIL loss is defined by
	\begin{equation} \label{eq:MIL_loss}
		\mathcal{L}_{MIL} = -y log(\bar{s})-(1-{y})log(1-\bar{s}),
	\end{equation}
	where $y\in\mathcal{Y}$ is the video-level label.
	
	\noindent\textbf{Inference.}
	In the inference stage, we comprehensively consider both the anomaly probabilities of the text descriptions within the video and the anomaly scores of the fused features. 
	Due to the potential inaccuracy of captions generated by the pre-trained model, directly adding the anomaly scores produced by the text modality and the fused modality could lead to misleading results. Therefore, we introduce the hyperparameter $\alpha$ as a balancing factor in the calculation of the final anomaly score, to adjust the relative weights of the anomaly scores derived from the text modality and the fused modality.
	Specifically, the anomaly score for each snippet can be calculated as follows:
	\begin{equation} \label{eq:final_anomaly_score}
		\hat{s}_i = \alpha s_i+ (1-\alpha)p_i,
	\end{equation}
	where $\alpha\in[0,1]$ is a hyperparameter for balancing two scores.
	
	\begin{table}[t]
		\caption{The AUC (\%) on UCF-Crime and AP (\%) on XD-Violence for model variants with varying designs in MSTA are provided.}
		\resizebox{.47\textwidth}{!}{
			\begin{tabular}{|cccc|cc|}
				\hline
				Index & Stage-I &  Stage-II  & Stage-III  &  UCF-Crime (\%)  &  XD-Violence (\%)  \\ 
				\hline
				1 & \XSolidBrush &  \XSolidBrush  &  \XSolidBrush   &  80.53 & 78.72\\
				2 & \Checkmark &  \XSolidBrush  &  \XSolidBrush    & 82.41 & 80.36\\
				3 & \Checkmark & \Checkmark     & \XSolidBrush    & 85.36 &  84.38 \\ 
				4 & \Checkmark & \Checkmark     & \Checkmark  & \textbf{89.67} & \textbf{85.92} \\ 
				\hline
			\end{tabular}
			\label{tab:ab_components_MSTA}
		}
	\end{table}
	
	\begin{table}[t]
		\caption{The AUC (\%) on UCF-Crime and AP (\%) on XD-Violence for model variants employing varied modalities are provided.}
		\resizebox{.47\textwidth}{!}{
			\begin{tabular}{|ccccc|cc|}
				\hline
				Index & Text &  RGB &  Flow  & Audio &  UCF-Crime  (\%)  &  XD-Violence  (\%)  \\ 
				\hline
				1 & \XSolidBrush &  \XSolidBrush  &  \Checkmark  &  \Checkmark & N/A  & 77.48 \\
				2 & \XSolidBrush &  \Checkmark  &  \Checkmark  &  \XSolidBrush & 87.46 & 80.82 \\
				3 & \XSolidBrush &  \Checkmark  &  \XSolidBrush  &  \Checkmark & N/A  & 84.12 \\
				4 & \XSolidBrush &  \Checkmark  &  \Checkmark  &  \Checkmark & N/A  & 84.78 \\
				\hline
				5 & \Checkmark &  \XSolidBrush  &  \XSolidBrush  &  \XSolidBrush & 79.86  & 70.21 \\
				\hline
				6 & \Checkmark &  \XSolidBrush  &  \Checkmark  &  \Checkmark & N/A  & 79.59 \\
				7 & \Checkmark &  \Checkmark  &  \Checkmark  &  \XSolidBrush & \textbf{89.67} & 82.75 \\
				8 & \Checkmark &  \Checkmark  &  \XSolidBrush  &  \Checkmark & N/A  & 84.69\\
				9 & \Checkmark &  \Checkmark  &  \Checkmark  &  \Checkmark & N/A & \textbf{85.92} \\
				\hline
			\end{tabular}
			\label{tab:ab_modalities}
		}
	\end{table}
	
	\section{Experiments}
	
	\subsection{Datasets and Evaluation Metrics}
	
	\subsubsection{Datasets}
	We evaluate our proposed framework on two large-scale public datasets, \ie UCF-Crime~\cite{sultani2018real} and XD-Violence~\cite{wu2020not}. The details of the datasets are as follows.
	
	\begin{itemize}	
		\item UCF-Crime~\cite{sultani2018real} is a large-scale dataset comprising 13 categories of anomalies, sourced from real-world indoor and outdoor surveillance footage, totaling 128 hours in video length. It contains 1,610 training videos and 290 test videos. Yuan~\etal~\cite{yuan2024towards} annotate the videos in UCF-Crime with captions, employing text to describe the events. We followed the annotations presented in this paper for our multimodal video anomaly detection task.
		
		\item XD-Violence~\cite{wu2020not} is another extensive dataset for video anomaly detection, featuring 4,754 untrimmed videos spanning 217 hours, gathered from online videos, real-life movies, surveillance footage, and more. The training set consists of 3,954 videos with video-level annotations, while the testing set includes 800 videos annotated at the frame level. Since this dataset lacks text annotations, we employed BLIP2~\cite{li2023blip} to generate text descriptions for video segments.
	\end{itemize}

	\subsubsection{Evaluation Metrics}
	To assess the performance of different models, we follow the common practice~\cite{sultani2018real,tian2021weakly,feng2021mist,URDMU_zh} by using the frame-level area under the ROC curve (AUC) as the evaluation metric for UCF-Crime, while employing the average precision (AP) for the XD-Violence dataset. A higher AUC or AP signifies better performance.
	
	\subsection{Implementation Details}
	The proposed framework is implemented in Pytorch and is trained on 4$\times$ NVIDIA Tesla V100 GPUs. In the MSTA, we employ the LLaMA-3-8B~\cite{meta2024introducing} as the LLM, the number of samplings $N_S$ is set to 10, the number of context samples $N_R$ is 80, and the threshold $\delta$ in Stage-III is set to 0.7. The number of generated samples is 17,000 for UCF-Crime and 65,000 for XD-Violence. In the unimodal encoders, the dimension of encoded features $D_E$ is 128, the number of attention heads $N_H$ is set to 4, and the Transformer layer $L_U$ is 1. In the multi-scale bottleneck Transformer, we use $L_M = 5$ fusion layers, and the number of bottleneck tokens in the first layer $N^{bt}_1$ is set to 16. In the global encoder, we set $L_G=3$ layers for the Transformer to aggregate global context. For the text extractor fine-tuning, we employ the bert-base-uncased as the base model, and we use the AdamW optimizer with a batch size of 256 and a learning rate of 0.00005 for 5 epochs. For the network training, the top-$K$ value in the MIL loss is employed by $K = 9$, the model is trained by using the SGD optimizer with a batch size of 32, and a learning rate of 0.0004 for UCF-Crime and 0.005 for XD-Violence. For the inference, the hyperparameter $\alpha$ is set to 0.5.	
	
	\begin{figure}[t]
		\centering
		\includegraphics[width=0.44\textwidth]{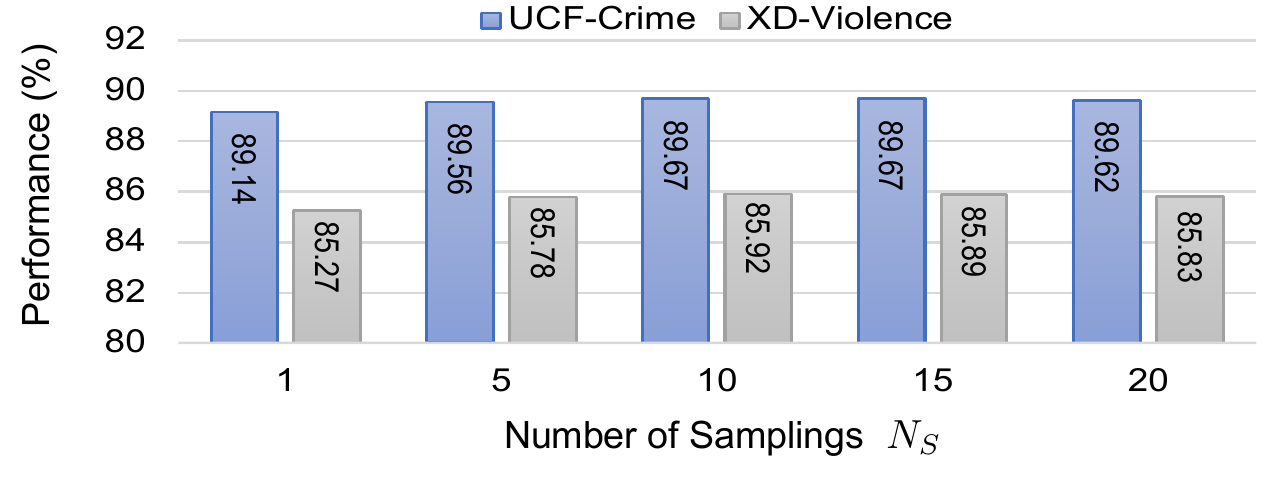} 
		\caption{The performance with a variant number of samplings $N_S$ in the Stage-II of MSTA. Best viewed in color.}	
		\label{fig:number_of_samplings}
		\vspace{-8pt}
	\end{figure}
	
	\subsection{Ablation Studies}
	
	\subsubsection{Effectiveness of The Proposed MSTA} \label{sec:ab_MSTA}
	We first conduct ablation experiments on the proposed MSTA to evaluate the effectiveness of multi-stage text augmentation. The experimental results are presented in Table~\ref{tab:ab_components_MSTA}. When all three stages (Stage-I, II, III) of the strategy are omitted (Index 1 in the table), relying solely on the extraction of text features using BERT, the performance of the model is suboptimal, evidenced by a 9.14\% decrease in AUC on the UCF-Crime dataset. The incorporation of caption summary (Index 2 in the table), which involves fine-tuning BERT with concise descriptive text, results in noticeable improvements in model performance, leading to increases of 1.88\% in AUC and 1.64\% in AP for the UCF-Crime and XD-Violence datasets, respectively. Further augmenting the process with pseudo-labeling of anomaly scores derived from video captions (Index 3 in the table) significantly enhances the sample size for fine-tuning BERT, which corresponds to a marked improvement in model performance, particularly observed as a 4.02\% increase in AP on the XD-Violence dataset. Finally, with the additional integration of text generation techniques (Index 4 in the table), which produces a substantial number of anomalous samples for fine-tuning the text feature extractor, the model achieves its best performance across both datasets.

	\begin{figure}[t]
		\centering
		\includegraphics[width=0.44\textwidth]{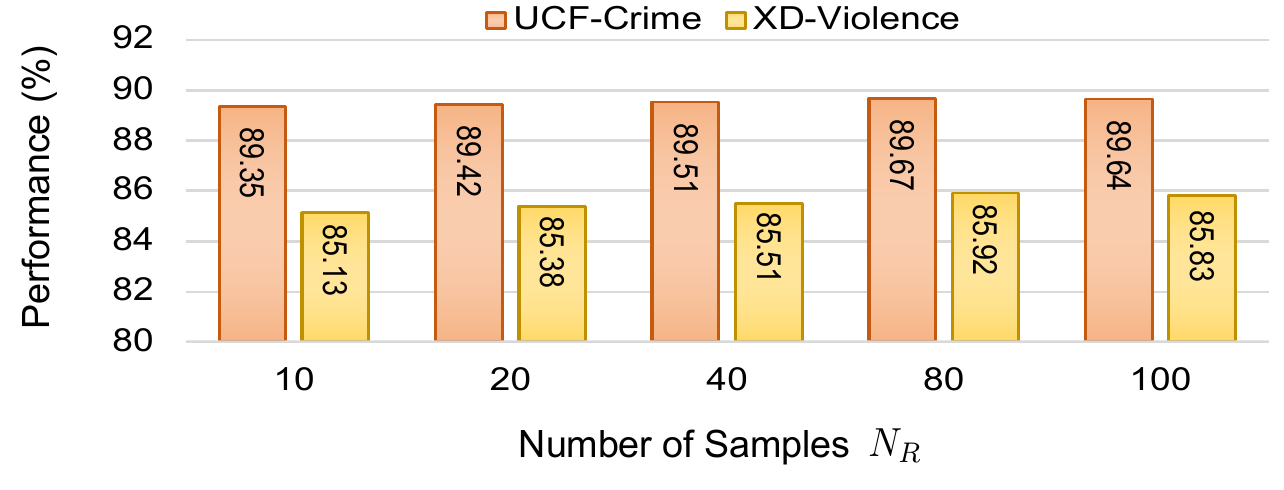} 
		\caption{The performance with a variant number of context samplings $N_R$ in the Stage-II(III) of MSTA. Best viewed in color.}	
		\label{fig:number_of_samples}
	\end{figure}
	
	\begin{figure}[t]
		\centering
		\includegraphics[width=0.46\textwidth]{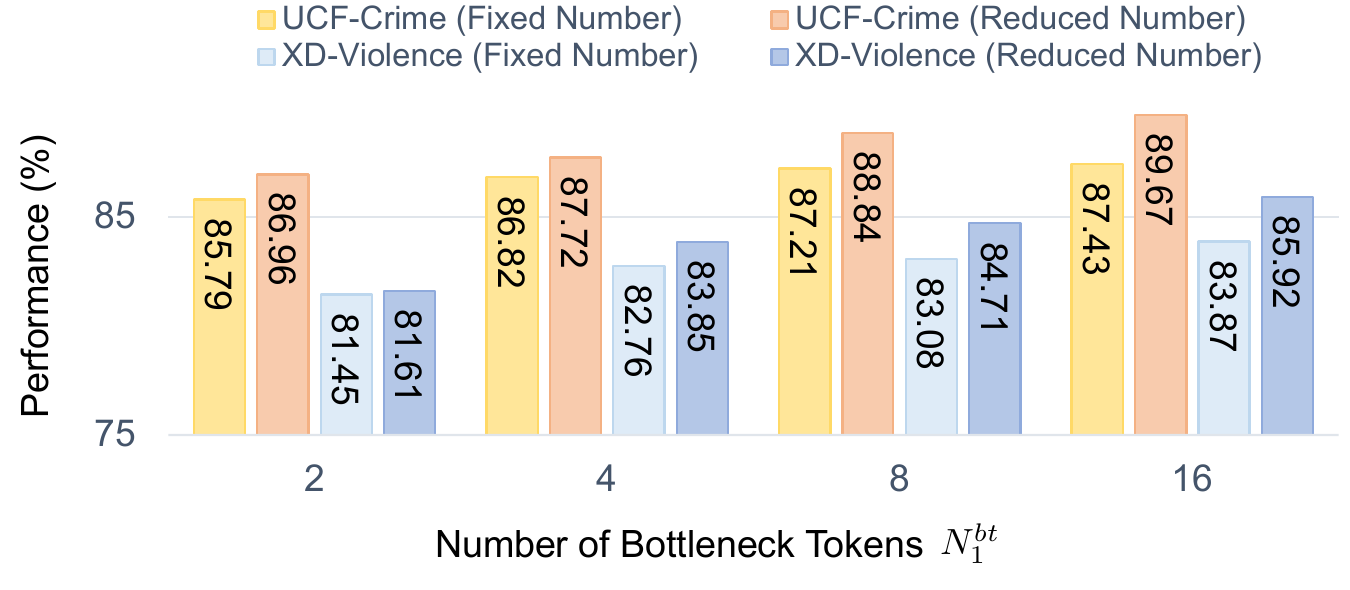} 
		\caption{The AUC (\%) on UCF-Crime and AP (\%) on XD-Violence are assessed for model variants utilizing a fixed or decreasing number of bottleneck tokens in the MSBT. The comparison of performance is conducted for the model across three modalities. The numbers 2, 4, 8, or 16 represent the number of tokens employed at the first layer (in the reducing case) or consistently across all layers (in the fixed case). Best viewed in color.}
		\label{fig:fixed_reduced_tokens}
	\end{figure}
	
	\begin{table}[t]
		\caption{The AUC (\%) on UCF-Crime and AP (\%) on XD-Violence for model variants with various designs in the multimodal fusion of MSBT are presented. In this context, 'CT' refers to the Cross-Transformer within the MSBT, 'T\_W' denotes the Transformer employed for weight computation, and 'Weighting' indicates the use of the weighting scheme.}
		\resizebox{.47\textwidth}{!}{
			\begin{tabular}{|cccc|cc|}
				\hline1
				Index & CT &  T\_W  & Weighting  &  UCF-Crime (\%)  &  XD-Violence (\%)  \\ 
				\hline
				1 & \XSolidBrush &  \XSolidBrush  &  \XSolidBrush   &  86.27 & 81.30 \\
				2 & \XSolidBrush &  \XSolidBrush  &  \Checkmark    & 86.53 & 81.71\\
				3 & \XSolidBrush & \Checkmark     & \Checkmark    & 87.04 &  82.06 \\ 
				\hline
				4 & \Checkmark & \XSolidBrush     & \XSolidBrush  &87.48 & 82.34 \\ 
				5 & \Checkmark & \XSolidBrush     & \Checkmark    & 87.62 &  82.69 \\ 
				6 & \Checkmark & \Checkmark     & \Checkmark    & \textbf{89.67} &  \textbf{85.92} \\ 
				\hline
			\end{tabular}
			\label{tab:ab_components_MSBT}
		}
	\end{table}
	
	\subsubsection{Effectiveness of the Text Guidance}
	We subsequently conduct ablation experiments to validate the effectiveness of introducing the text modality for guidance, the experimental results are summarized in Table~\ref{tab:ab_modalities}. Given the absence of the audio modality in the UCF-Crime dataset, our experiments utilize a maximum of three modalities: text, RGB, and flow. The findings reveal that by employing our proposed MSBT module, the model achieved an AUC of 87.46\% and an AP of 80.82\% on the UCF-Crime and XD-Violence datasets, respectively (Index 2 in the table). The introduction of the text modality further enhances the model's performance on UCF-Crime and XD-Violence, leading to improvements of 2.21\% in AUC and 1.93\% in AP, respectively. We also report the results of the model utilizing solely the text modality (Index 5 in the table), as determined by the anomaly probabilities produced in Eq.~(\ref{eq:text_anomaly_score}). Our findings indicate that the model's performance when relying solely on the text modality is unsatisfactory. Furthermore, it is evident that the addition of the text modality consistently improves the model's performance, with the highest overall performance attained when all modalities are utilized, \eg an AP of 85.92\% was achieved on the XD-Violence dataset.

	\subsubsection{Impact of the Number of Samplings $N_S$}
	To investigate the impact of the number of sampling $N_S$ in Stage-II of the MSTA on model performance, we conduct experiments varying $N_S$ from 1 to 20, and the experimental results are illustrated in Fig.~\ref{fig:number_of_samplings}. From the results, we can observe that the model achieves the best performance on the UCF-Crime dataset with $N_S$ values of 10 or 15, while the best performance on the XD-Violence dataset occurs at $N_S = 10$. Consequently, we set the $N_S$ to 10 for our experiments. Additionally, we note that changes in $N_S$ have minimal effect on model performance, \eg the AUC on the UCF-Crime dataset ranged from 89.14\% to 89.62\%. This suggests that the model exhibits a degree of insensitivity to variations in $N_S$ during the training process.
	
	\begin{figure}[t]
		\centering
		\includegraphics[width=0.46\textwidth]{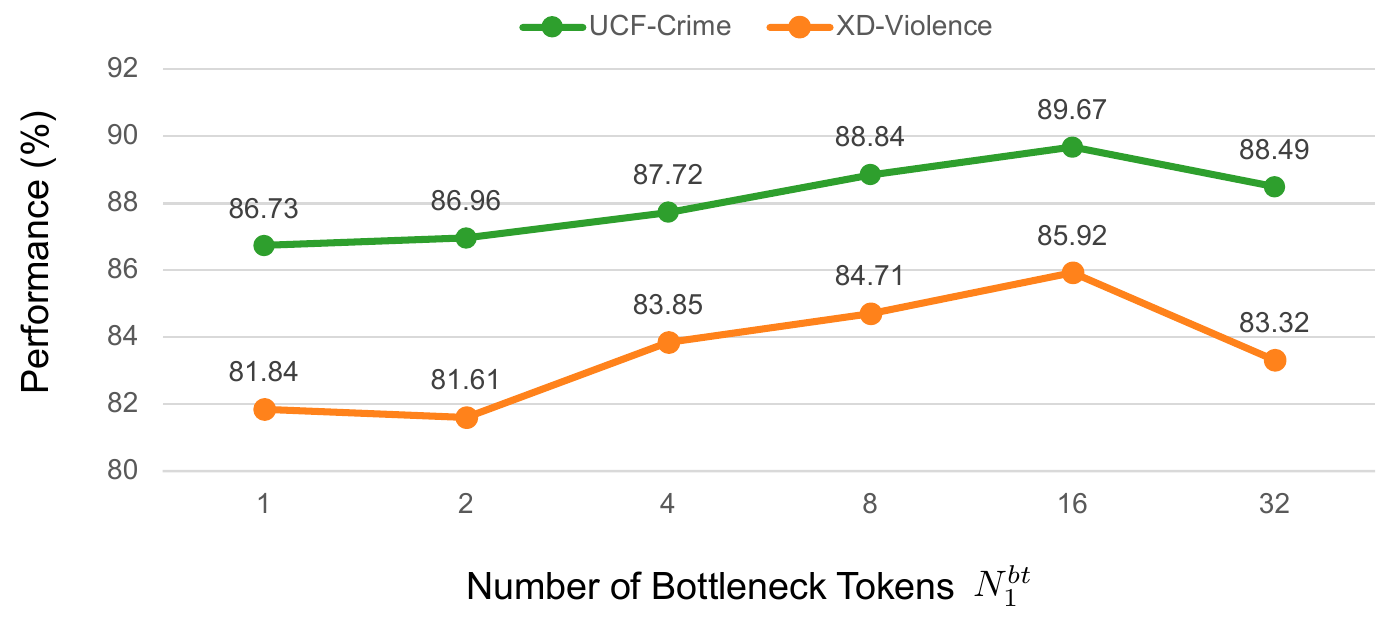} 
		\caption{The performance evaluation was conducted with varying token numbers in the initial layer of our multi-scale bottleneck Transformer. Best viewed in color.}	
		\label{fig:number_of_bottleneck}
	\end{figure}
	
	\subsubsection{Impact of the Number of Context Samples $N_R$}
	To further investigate the influence of the number of context samples $N_R$ in Stages-II and Stage-III of the MSTA on model performance, we conduct experiments by changing $N_R $ from 10 to 100. The results of these experiments are illustrated in Fig.~\ref{fig:number_of_samples}. Notably, the model attains the best results on both datasets when $N_R$ is set to 80. Additionally, we note that variations in $N_R$ result in minimal changes to the model's performance, with fluctuation ranges of 0.32\% and 0.79\% observed on the UCF-Crime and XD-Violence datasets, respectively.
	
	\subsubsection{Effectiveness of the Designs in Multimodal Fusion}
	We then delve into the multimodal fusion module, which comprises the multi-scale bottleneck Transformer (MSBT) and a bottleneck token-based weighting sub-module. Each component is validated through comparisons with various alternative designs.
	
	In the MSBT, our current design incorporates two key features. First, the bottleneck tokens from each layer are derived from those learned in the previous layer via the Cross-Transformer (CT) defined in Eq.~(\ref{eq:MSBT_CT}), facilitating a gradual condensation of information. An alternative approach would be to skip the Cross-Transformer, resulting in tokens that are randomly initialized at the start of each layer with no information carried over from preceding layers. As shown in Table~\ref{tab:ab_components_MSBT}, our findings demonstrate the superiority of using the CT. Second, the number of bottleneck tokens in consecutive layers of our current model is halved to promote gradual condensation. In contrast, one could maintain a fixed number of tokens, as seen in other methods~\cite{nagrani2021attention,Yan2022Transformer}. Fig.~\ref{fig:fixed_reduced_tokens} provides a comparison, validating the effectiveness of our gradual condensing strategy.

	\begin{figure}[t]
		\centering
		\includegraphics[width=0.45\textwidth]{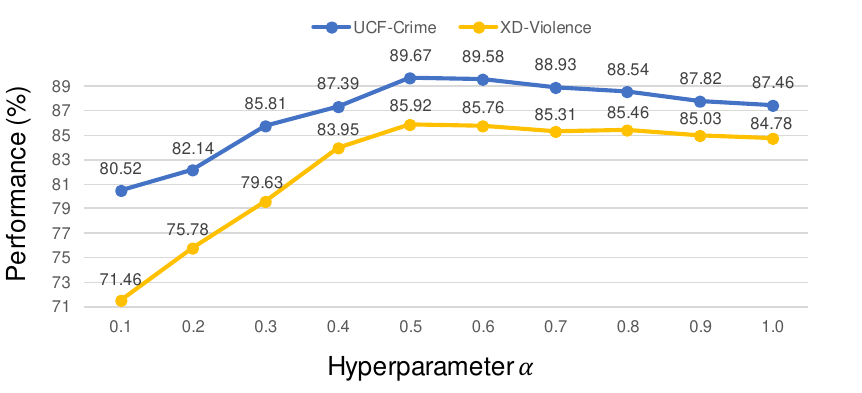} 
		\caption{The performance of the model on two datasets with different hyperparameter settings for $\alpha$. Best viewed in color.}	
		\label{fig:hyperparameter_alpha}
	\end{figure}

	\begin{figure}[t]
		\centering
		\includegraphics[width=0.45\textwidth]{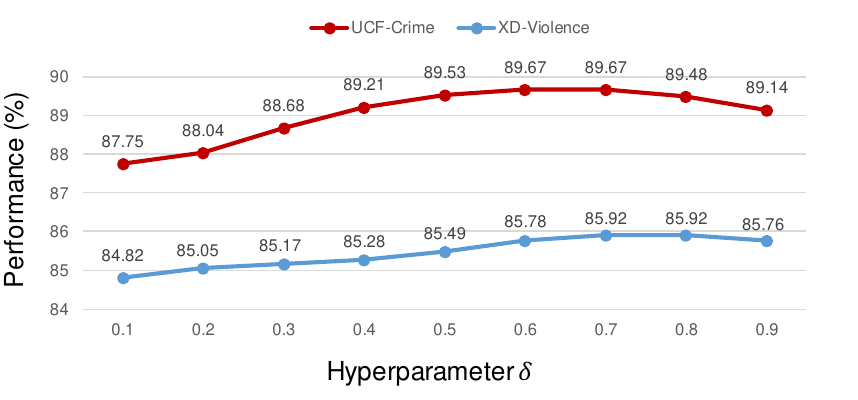} 
		\caption{The performance of the model on two datasets with different hyperparameter settings for $\delta$. Best viewed in color.}	
		\label{fig:hyperparameter_delta}
		\vspace{-8pt}
	\end{figure}

	\begin{table}[t]
		\centering
		\caption{Frame-level AUC (\%) on UCF-Crime dataset of various state-of-the-art methods. The best results are \textbf{bold}.
		}
		\resizebox{0.47\textwidth}{!}{
			\begin{tabular}{|l|c|c|c|}
				\hline
				Method & Venue & Modality &  AUC (\%)\\
				\hline
				Sultani~\etal~\cite{sultani2018real} &   CVPR'18 & RGB &  75.41  \\
				Wu~\etal~\cite{wu2021learning} &  TIP'21  & RGB  &  84.89    \\
				RTFM~\cite{tian2021weakly}     &  ICCV'21   & RGB  &   84.30      \\
				MIST~\cite{feng2021mist}     &  CVPR'21   & RGB  &   82.30      \\
				DAR~\cite{liu2022decouple}& TIFS'23  & RGB   &  85.18    \\
				LSTC~\cite{sun2023long} & ICME'23  & RGB   &  85.88    \\
				UR-DMU~\cite{URDMU_zh}& AAAI'23  & RGB   &  86.97   \\
				MSAF~\cite{wei2022msaf}& SPL'22  & RGB+Flow  & 81.34   \\
				DAR~\cite{liu2022decouple}& TIFS'23  & RGB+Flow  & 85.36   \\
				Tan~\etal~\cite{tan2024overlooked} & WACVW'24  & RGB+Flow  & 86.71   \\
				TEVAD~\cite{chen2023tevad} & CVPRW'23  & Text+RGB  & 84.90   \\
				Shi~\etal~\cite{shi2024caption} & Access'24  & Text+RGB  & 84.64   \\
				Yuan~\etal~\cite{yuan2024surveillance} &TCSVT'24 & Text+RGB & 85.30 \\
				\hline
				TG-MVAD (Ours) & This Work & RGB+Flow & \textbf{87.46} \\
				TG-MVAD (Ours) & This Work & Text+RGB & \textbf{89.18} \\
				TG-MVAD (Ours) & This Work & Text+RGB+Flow & \textbf{89.67} \\
				
				\hline
			\end{tabular}
		}
		\label{tab:UCF_results}
	\end{table}
	
	Regarding the bottleneck token-based weighting sub-module, we evaluate three alternative designs. The first design leaves out the Transformer employed in Eq.~(\ref{eq:MSBT_W}), while the second eliminates the weighting. Results presented in Table~\ref{tab:ab_components_MSBT} indicate that the weighting scheme enhances performance, highlighting the significant impact of the weight computation method on outcomes.

	\subsubsection{Impact of the Number of Bottleneck Token $N^{bt}_1$}
	Bottleneck tokens play a crucial role in the MSBT module. To investigate how the quantity of bottleneck tokens affects performance, we conduct experiments by varying the number of tokens $N^{bt}_1$ from 1 to 32. This range corresponds to the employment of 1 to 5 MSBT layers following a gradual reduction scheme. The results are illustrated in Fig.~\ref{fig:number_of_bottleneck}. The results presented in the figure indicate that as the number of bottleneck tokens increases, the model's performance exhibits an upward trend. Notably, the model achieves the best performance on both datasets when the number of bottleneck tokens is set to 16 in the first layer.
	
	\subsubsection{Impact of the Hyperparameter $\alpha$}
	The hyperparameter $\alpha$ is used to balance the anomaly probabilities derived from textual descriptions and fused modality when calculating anomaly scores during inference. To further investigate the impact of $\alpha$ on model performance, we conducted experiments where $\alpha$ was varied from 0.1 to 1.0, with the results shown in Fig.~\ref{fig:hyperparameter_alpha}. The experimental results indicate that as $\alpha$ increases from 0.1 to 0.5, with the weight on the fused modality-based anomaly scores progressively increasing, the model's performance improves continuously. Notably, when $\alpha$ is set to 0.5, the model achieves the best performance on both datasets. However, as $\alpha$ increases from 0.5 to 1.0, with the weight on the text modality-based anomaly scores gradually decreasing, the model's performance begins to decline. In summary, while the hyperparameter $\alpha$ does influence model performance, the best results are obtained when $\alpha$ is set to 0.5 for both datasets. Therefore, $\alpha$ can be set to a value of 0.5 for inference.
	
	\subsubsection{Impact of the Hyperparameter $\delta$}
	The hyperparameter $\delta$ serves as a threshold in Stage-III to determine whether the generated text describes an anomalous event, with texts having an anomaly score greater than $\delta$ being classified as descriptions of anomalous events. We conducted experiments to investigate the effect of $\delta$ on model performance, varying $\delta$ from 0.1 to 0.9, with the results presented in Fig.~\ref{fig:hyperparameter_delta}. The experimental results show that as $\delta$ increases, the model's performance initially improves and then slowly declines. Notably, when $\delta$ is set to 0.6 or 0.7, the model achieves the best performance on the UCF-Crime dataset, while when $\delta$ is set to 0.7 or 0.8, the model performs best on the XD-Violence dataset. Considering these results, setting $\delta$ to 0.7 is optimal, as it enables the model to achieve the best performance on both datasets.
	
	\begin{figure*}[t]
		\centering
		\includegraphics[width=0.97\textwidth]{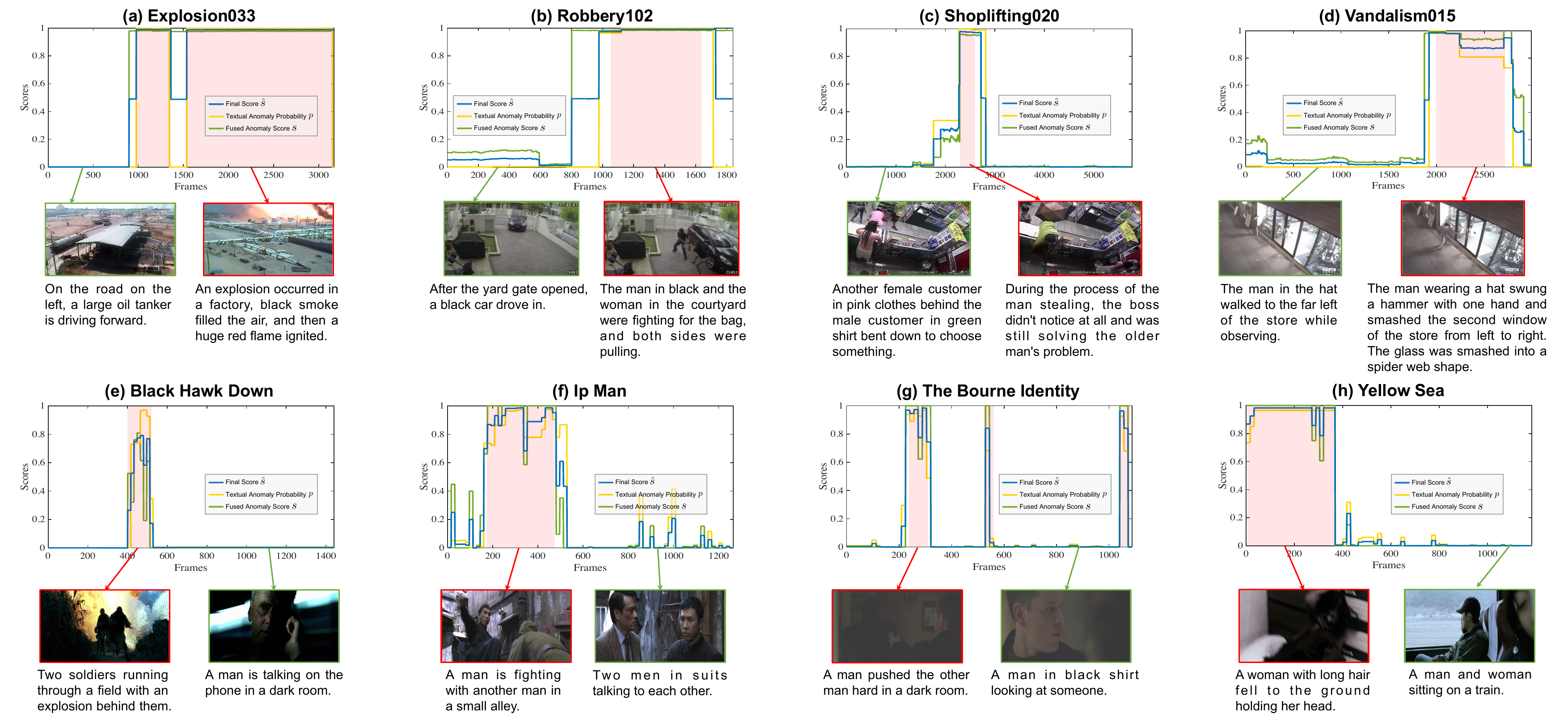} 
		\caption{Visualization of anomaly scores predicted on the (a-d) UCF-Crime and (e-h) XD-Violence test sets. The pink regions indicate ground-truth abnormal events. Best viewed in color.}	
		\label{fig:anomaly_scores}
		\vspace{-10pt}
	\end{figure*}

	\subsection{Comparison to State-of-the-Art}
	
	\subsubsection{Results on UCF-Crime Dataset}
	The frame-level AUC performance of various state-of-the-art (SOTA) methods on the UCF-Crime dataset is summarized in Table~\ref{tab:UCF_results}. Notably, traditional RGB-based approaches like Sultani~\etal~\cite{sultani2018real} and MIST~\cite{feng2021mist} demonstrate lower AUC of 75.41\% and 82.30\%, respectively. In contrast, the inclusion of flow modality enhances performance significantly, as seen with methods such as DAR~\cite{liu2022decouple} and Tan and Liu~\cite{tan2024overlooked}, which achieve AUC of 85.36\% and 86.71\% for RGB+Flow modalities. 
	
	Comparatively, our proposed method stands out with substantial improvements. Utilizing RGB+Flow achieves an AUC of 87.46\%, while the combination of Text+RGB and Text+RGB+Flow further enhances the results to 89.18\% and a remarkable 89.67\%, respectively. In addition, compared to the TEVAD~\cite{chen2023tevad}, which similarly employs text modalities, our proposed approach demonstrates an improvement of 4.28\% in AUC when utilizing both text and RGB modalities. These results indicate that our approach not only surpasses existing methods in terms of accuracy but also emphasizes the effectiveness of integrating text information with visual modalities to enhance performance on the UCF-Crime dataset.
	
	\begin{table}[t]
		\centering
		\caption{Frame-level AP (\%) on XD-Violence dataset of various state-of-the-art methods. The best results are \textbf{bold}.
		}
		\resizebox{0.47\textwidth}{!}{
			\begin{tabular}{|l|c|c|c|}
				\hline
				Method & Venue & Modality   & AP (\%) \\
				\hline
				Wu \etal~\cite{wu2021learning} &  TIP'21  & RGB   & 75.90      \\
				RTFM~\cite{tian2021weakly}     &  ICCV'21   & RGB    & 77.81     \\
				DAR~\cite{liu2022decouple}& TIFS'23  & RGB   &    78.94  \\
				UR-DMU~\cite{URDMU_zh} & AAAI'23  & RGB   &    81.66  \\
				HL-Net~\cite{wu2020not}       &  ECCV'20  & RGB+Audio  &  78.64      \\
				Pang~\etal~\cite{pang2021violence}& ICASSP'21  & RGB+Audio  & 81.69       \\
				ACF~\cite{wei2022look}& ICASSP'22  & RGB+Audio   & 80.13       \\  
				Pang et al.~\cite{pang2022audiovisual}& TMM'22  & RGB+Audio   & 79.37         \\
				MACIL-SD~\cite{yu2022modality} & MM'22  & RGB+Audio & 83.40         \\
				Wu~\etal~\cite{wu2022weakly}      & TMM'22    & RGB+Audio  & 78.64         \\
				MSAF~\cite{wei2022msaf}& SPL'22  & RGB+Audio  & 80.51   \\
				DAR~\cite{liu2022decouple}& TIFS'23  & RGB+Audio  & 79.32         \\
				Zhang~\etal~\cite{zhang2023exploiting}      & CVPR'23    & RGB+Audio   & 81.43     \\
				DAR~\cite{liu2022decouple}& TIFS'23  & RGB+Flow  & 79.15         \\
				Wu~\etal~\cite{wu2022weakly}      & TMM'22    & Audio+Flow   &  72.96         \\ 
				TEVAD~\cite{chen2023tevad}  & CVPRW'23  & Text+RGB  & 79.80   \\
				Wu~\etal~\cite{wu2022weakly}      & TMM'22    & RGB+Audio+Flow  & 79.53        \\
				Xiao~\etal~\cite{xiao2023scoreformer}      & ICASSP'23    & RGB+Audio+Flow  & 84.54     \\
				Na~\etal~\cite{na2024leveraging}      & MLKE'24    & Text+RGB+Audio+Flow & 83.90     \\
				\hline
				TG-MVAD (Ours) & This Work & RGB+Audio &  \textbf{84.12} \\
				TG-MVAD (Ours) & This Work & RGB+Flow &  \textbf{80.82} \\
				TG-MVAD (Ours) & This Work & Audio+Flow &  \textbf{77.48} \\
				TG-MVAD (Ours) & This Work & Text+RGB &  \textbf{83.75} \\
				TG-MVAD (Ours) & This Work & RGB+Audio+Flow &  \textbf{84.78} \\
				TG-MVAD (Ours) & This Work & Text+RGB+Audio+Flow &  \textbf{85.92} \\
				
				\hline
			\end{tabular}
			
		}
		\vspace{-8pt}
		\label{tab:XD_results}
	\end{table}
	
	\subsubsection{Results on XD-Violence Dataset}
	Table~\ref{tab:XD_results} presents the frame-level AP performance of various SOTA methods on the XD-Violence dataset. The results indicate a significant performance range across different modalities, with many methods achieving competitive results in RGB and RGB+Audio modalities. For instance, methods like UR-DMU~\cite{URDMU_zh} and MACIL-SD~\cite{yu2022modality} yield AP of 81.66\% and 83.40\%, respectively, using the RGB+Audio modality, which highlights the efficacy of audio modality in enhancing performance. In comparison, our proposed method outperforms several existing approaches, achieving notable AP  in various configurations, including RGB+Audio (84.12\%), RGB+Flow (80.82\%), and a maximum of 85.92\% with the combined modality of text, RGB, audio, and flow. This suggests that the integration of multiple modalities significantly contributes to improving detection capabilities, affirming the effectiveness of our approach in addressing the XD-Violence dataset. 
	
	\begin{figure*}[t]
		\centering
		\includegraphics[width=0.98\textwidth]{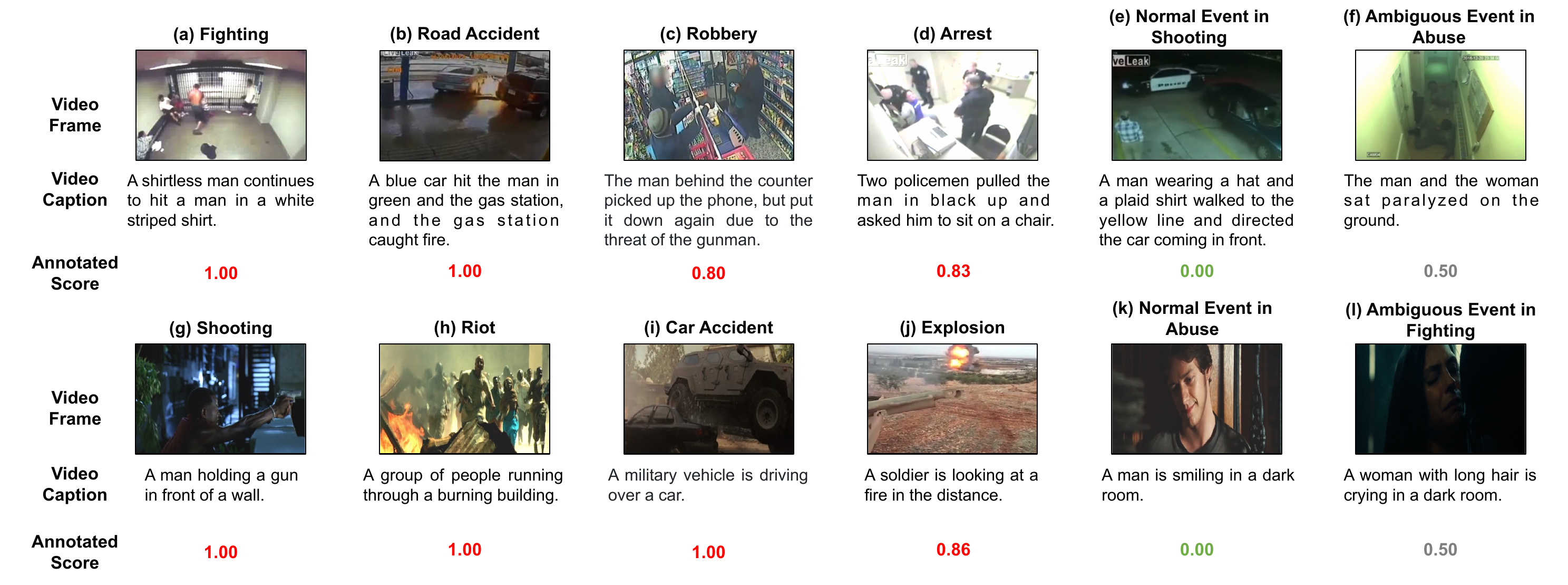} 
		\caption{The visualization of score annotation for video segments in the (a-f) UCF-Crime and (g-l) XD-Violence datasets was conducted utilizing the Stage-II labeling strategy.}	
		\label{fig:visual_stageII}
		\vspace{-10pt}
	\end{figure*}

	\subsection{Qualitative Analyses}
	
	\subsubsection{Predicted Anomaly Scores}
	We first visualize the anomaly scores predicted by our proposed method on the test sets of the UCF-Crime and XD-Violence datasets, as shown in Fig.~\ref{fig:anomaly_scores}. To better illustrate the impact of incorporating text anomaly probabilities on the model's performance during inference, we plot three distinct anomaly score curves for each showcased video. The yellow line represents the text anomaly probability (as per Eq.~(\ref{eq:text_anomaly_score})), the green line denotes the fused feature's anomaly score (according to Eq.~(\ref{eq:fused_anomaly_score})), and the blue line reflects the final weighted anomaly score (as indicated in Eq.~(\ref{eq:final_anomaly_score})). From the figure, we can observe that the text anomaly probability and the fused anomaly score complement each other effectively. For instance, in the short-term normal segment of \textit{(a) Explosion033} (frames 1361$\sim$1536), the fused anomaly score $s$ is relatively high, while the lower text anomaly probability $p$ reduces the final weighted anomaly score $\hat{s}$, thereby avoiding false alarms. Similarly, in the normal segment of \textit{(b) Robbery102} (frames 801$\sim$976), the text anomaly probability also contributes to lowering the final anomaly score.
	
	Moreover, the fused anomaly score aids in mitigating false alarms as demonstrated in \textit{(f) Ip Man}, specifically between frames 849$\sim$864 and 977$\sim$1008. Additionally, the text anomaly probability proves beneficial for detecting anomalies. For instance, in the anomalous segment of \textit{(e) Black Hawk Down} (frames 465$\sim$480), the text anomaly probability is recorded at 0.97, while the fused anomaly score stands at 0.61. This combination elevates the final anomaly score to 0.79 after weighting.
	
	Furthermore, it is noteworthy that our proposed model is capable of detecting not only individual anomalous events but also accurately identifying multiple anomalous events within a single video, as exemplified by the three anomalous events present in \textit{(g) The Bourne Identity}.

	\subsubsection{Annotated Scores in Stage-II}
	To better illustrate the effect of the annotation strategy employed in Stage-II for scoring video segments, selected annotated segments from the UCF-Crime and XD-Violence datasets have been presented in Fig.~\ref{fig:visual_stageII}. The annotated results presented indicate that the label predictions obtained through Stage-II effectively assign accurate pseudo-labels to certain video segments. For instance, abnormal segments in the UCF-Crime dataset, such as \textit{(a) Fighting} and \textit{(b) Road Accident}, as well as \textit{(g) Shooting} and \textit{(h) Riot} from the XD-Violence dataset, have all been assigned an anomaly score of 1.00. Similarly, normal events within the abnormal videos, exemplified by \textit{(e) Normal Event in Shooting} or \textit{(k) Normal Event in Abuse}, have been correctly labeled with an anomaly score of 0.00.
	
	Moreover, it was observed that some ambiguous event descriptions in the videos pose challenges in determining whether they constitute anomalies. For instance, the descriptions depicted in \textit{(f) Ambiguous Event in Abuse} and \textit{(l) Ambiguous Event in Fighting} were both assigned pseudo-labels with an anomaly score of 0.50. This underscores the complexities involved in accurately categorizing events that lack clear distinguishing characteristics.

\begin{table}[t]
	\centering
	\caption{The AP performance (\%) on the XD-Violence dataset, model params, and floating point operations (FLOPs) of the proposed MSBT-based video anomaly detection framework are compared with other methods. Here, ``R'', ``F'', ``A'', and ``T'' correspond to the RGB, optical flow, audio, and text modalities, respectively. $\dagger$ indicates that this result is referenced from the work of Yu~\textit{et al.}~\cite{yu2022modality}, and * denotes results obtained by retraining the official code with different input modality features.}
	\resizebox{.48\textwidth}{!}{
		\begin{tabular}{|c|c|c|c|c|}
			\hline
			Method & Modality  & AP (\%)  & Params  & FLOPs     \\ 
			\hline
			MACIL-SD(Light)~\cite{yu2022modality} & R+A & 82.17 & 0.347M & 29.2G \\
			MACIL-SD(Full)~\cite{yu2022modality} & R+A & 83.40 & 0.678M & 56.4G \\
			MACIL-SD(Full)~\cite{yu2022modality}* & R+F & 79.73 & 0.792M & 65.6G \\
			RTFM~\cite{tian2021weakly}$\dagger$ & R+A & 78.54 & 13.190M & 1013.6 G \\
			Wu~\textit{et al.}~\cite{wu2020not}$\dagger$ & R+A & 78.66 & 1.539M & N/A \\
			Pang~\textit{et al.}~\cite{pang2021violence} & R+A & 81.69 & 1.876M & N/A\\
			\hline
			\multirow{3}{*}{This work} & \multirow{3}{*}{R+A} & \multirow{3}{*}{84.12} & 0.685M  & 25.8G \\
			&  &  & (0.154M (MSBT) &  (10.4G (MSBT) \\
			&  &  & + 0.531M (Others)& + 15.4G (Others))\\
			\hline
			\multirow{3}{*}{This work} & \multirow{3}{*}{R+F} & \multirow{3}{*}{80.82} & 0.743M  & 30.8G  \\
			&  &  & (0.154M (MSBT) & (11.4G (MSBT)\\
			&  &  & 0.589M (Others)) & + 19.4G (Others))\\
			\hline
			\multirow{3}{*}{This work} & \multirow{3}{*}{R+A+F} & \multirow{3}{*}{84.78} & 3.252M  & 102.2G  \\
			&  &  & (0.154M (MSBT) & (25.6G (MSBT)\\
			&  &  & + 3.098M (Others)) & + 76.6G (Others))\\
			\hline
			\multirow{3}{*}{This work} & \multirow{3}{*}{T+R+A+F} & \multirow{3}{*}{85.92} & 11.478M  & 288.2G  \\
			&  &  & (0.155M (MSBT) & (50.0G (MSBT)\\
			&  &  & + 11.323M (Others)) & + 238.2G (Others))\\
			\hline
		\end{tabular}
		\label{tab:complexity}
	}
\end{table}
	
	\subsection{Complexity Analysis}
	We further analyze the number of parameters and floating-point operations (FLOPs) of the proposed MSBT-based multimodal video anomaly detection framework and compare these results with those of other methods. The results are shown in Table~\ref{tab:complexity}. As indicated in the table, we divide the number of parameters and FLOPs of the MSBT-based anomaly detection framework into two parts: the former refers to the parameter count of the feature fusion module based on the multi-scale bottleneck Transformer (MSBT), while the latter includes the parameters of other modules, such as the unimodal encoders and global encoder. The MSBT-based feature fusion module can be regarded as a plug-and-play module, meaning it can be integrated into other baseline networks. 
	Taking the case of using both RGB visual and audio modalities, the parameter count of our proposed method is comparable to that of the MACIL-SD (Full) method proposed by Yu~\textit{et al.}, yet its AP performance improves by 0.72\%. Furthermore, our method requires fewer parameters than those of Wu~\textit{et al.}~\cite{wu2020not} and Pang~\textit{et al.}~\cite{pang2021violence}, while outperforming both in terms of performance. Additionally, we analyze the efficiency of the proposed multi-stage text augmentation (MSTA) method. In this paper, we use the LLaMA-3-8B model for inference in in-context learning. In Stage-I, the average time for summarizing one video is 0.93 seconds. In Stage-II, with the number of sampling $N_s=10$ and the number of context samples $N_R=80$, the average time to label a single sample is 3.58 seconds. In Stage-III, when the number of context samples is also set to 80, the average time to generate a new sample is 2.65 seconds.
	
	\section{Conclusion}
	In this paper, we propose a novel multi-stage text augmentation (MSTA) mechanism that successfully mitigates training bias by generating high-quality abnormal text samples. Our approach enriches the training dataset with diverse, informative text descriptions, thereby reducing the imbalance between normal and abnormal samples and promoting a more effective extractor fine-tuning. Complementarily, we introduce a feature fusion module based on a multi-scale bottleneck Transformer (MSBT), which enhances inter-modality integration by utilizing a streamlined set of bottleneck tokens. This innovative architecture captures complex dependencies across modalities. The experimental results obtained from large-scale datasets, UCF-Crime and XD-Violence, illustrate that our methodology achieves SOTA performance across multiple modalities, including text, RGB, optical flow, and audio. Notably, ablation studies reveal that the proposed MSTA significantly validates the model's effectiveness. Overall, our findings underscore the importance of robust text augmentation and efficient multimodal integration as crucial components in advancing model performance in anomaly detection tasks.
	
	\section*{Acknowledgments}
	This research was funded by the Zhejiang Province Pioneer Research and Development Project “Research on Multi-modal Traffic Accident Holographic Restoration and Scene Database Construction Based on Vehicle-cloud Intersection” (Grant No. 2024C01017).

	\ifCLASSOPTIONcaptionsoff
	\newpage
	\fi

	\bibliographystyle{IEEEtran}
	\bibliography{IEEEfull}
	
\end{document}